\documentclass[a4paper,11pt]{article}

\usepackage{amsmath,amssymb,amsthm,bbm,bm}  
\usepackage{authblk}  
\usepackage[british]{babel}
\usepackage{balance}
\usepackage[sorting=none,doi=false,isbn=false,giveninits=true,style=nature,url=false]{biblatex}           
\usepackage[T1]{fontenc}
\usepackage[top=2.5cm, bottom=2.5cm, left=2.5cm, right=2.5cm]{geometry}
\usepackage{graphbox}       

\usepackage[utf8]{inputenc}
\usepackage[mono=false]{libertine}  

\usepackage{csquotes}  
\usepackage{booktabs} 
\usepackage{caption}
\usepackage{color}
\usepackage{comment}
\usepackage{enumitem}
\usepackage{graphicx}
\PassOptionsToPackage{hyphens}{url}  
\usepackage[colorlinks = true,
            linkcolor = blue,
            urlcolor  = blue,
            citecolor = blue,
            anchorcolor = blue]{hyperref}
\usepackage{cleveref}  

\usepackage{mathtools}
\usepackage{microtype}      
\usepackage{nicefrac}
\usepackage{physics}
\usepackage[usenames,dvipsnames]{xcolor}
\usepackage[most]{tcolorbox}
\usepackage{subfig}
\DeclareUnicodeCharacter{1EF3}{{\'y}}

\newcommand{\R}{\mathbb{R}}
\newcommand{\N}{\mathbb{N}}
\newcommand{\Q}{\mathcal{Q}}
\newcommand{\bQ}{\boldsymbol{\Q}}
\renewcommand{\S}{\mathbb{S}}

\newcommand{\normal}{\mathcal{N}}
\newcommand{\EE}{\mathbb{E}\,}

\usepackage{xcolor}
\definecolor{C0}{HTML}{1f77b4}
\definecolor{C1}{HTML}{ff7f0e}
\definecolor{C2}{HTML}{2ca02c}
\definecolor{C3}{HTML}{d62728}
\definecolor{C4}{HTML}{9467bd}
\definecolor{C5}{HTML}{8c564b}
\definecolor{C6}{HTML}{e377c2}
\definecolor{C7}{HTML}{7f7f7f}
\definecolor{C8}{HTML}{bcbd22}
\definecolor{C9}{HTML}{17becf}

\renewcommand{\th}{\mathrm{th}}

\newcommand{\gs}{\mathrm{gs}}
\newcommand{\tjump}{t_\mathrm{cross}}
\newcommand{\opt}{\mathrm{opt}}
\newcommand{\loss}{\mathcal{L}}
\newcommand{\spherical}{z}
\newcommand{\noise}{\xi}
\newcommand{\bnoise}{\boldsymbol{\xi}}
\newcommand{\coupling}{J}
\newcommand{\lr}{\eta}
\newcommand{\spin}{x}

\newcommand{\lrexponent}{\beta}

\newcommand{\signal}{x^\star}
\newcommand{\ctti}{C_{\mathrm{TTI}}}
\newcommand{\rtti}{R_{\mathrm{TTI}}}
\newcommand{\ca}{\mathcal{C}}
\newcommand{\ra}{\mathcal{R}}
\newcommand{\muinf}{\spherical_{\infty}}
\newcommand{\mse}{\mathrm{mse}}
\newcommand{\bs}{B}
\newcommand{\tsim}{\!~\sim~\!}
\newcommand{\agingexp}{\gamma}
\newcommand{\hessian}{\mathrm{Hess}}
\let\oldto\to  
\renewcommand{\to}{{\oldto}} 

\newcommand{\citet}[1]{\textcite{#1}}
\bibliography{refs}

\title{Optimal learning rate schedules in high-dimensional non-convex optimization problems}
\author[1, 2]{St\'ephane d'Ascoli$^{\star\dagger}$}
\author[1, 3]{Maria Refinetti$^{\star\ddagger}$}
\author[1]{Giulio Biroli      }

\affil[1]{Department of Physics, \'Ecole Normale Sup\'erieure, Paris,
  France}
\affil[2]{Meta AI, Paris, France}
\affil[3]{IdePHICS Laboratory, EPFL, Switzerland}

\date{}

\begin{document}

\def\thefootnote{$\star$}\footnotetext{Equal contribution.}\def\thefootnote{\arabic{footnote}}
\def\thefootnote{$\dagger$}\footnotetext{stephane.dascoli@gmail.com}\def\thefootnote{\arabic{footnote}}
\def\thefootnote{$\ddagger$}\footnotetext{mariaref@gmail.com}\def\thefootnote{\arabic{footnote}}

\maketitle
\begin{abstract}
Learning rate schedules are ubiquitously used to speed up and improve optimisation. Many different policies have been introduced on an empirical basis, and theoretical analyses have been developed for convex settings. However, in many realistic problems the loss-landscape is high-dimensional and non convex -- a case for which results are scarce. In this paper we present a first analytical study of the role of learning rate scheduling in this setting, focusing on Langevin optimization with a learning rate decaying as $\eta(t)=t^{-\beta}$. We begin by considering models where the loss is a Gaussian random function on the $N$-dimensional sphere ($N\rightarrow \infty$), featuring an extensive number of critical points. We find that to speed up optimization without getting stuck in saddles, one must choose a decay rate $\beta<1$, contrary to convex setups where $\beta=1$ is generally optimal. We then add to the problem a signal to be recovered. In this setting, the dynamics decompose into two phases: an \emph{exploration} phase where the dynamics navigates through rough parts of the landscape, followed by a \emph{convergence} phase where the signal is detected and the dynamics enter a convex basin. In this case, it is optimal to keep a large learning rate during the exploration phase to escape the non-convex region as quickly as possible, then use the convex criterion $\beta=1$ to converge rapidly to the solution. Finally, we demonstrate that our conclusions hold in a common regression task involving neural networks.
\end{abstract}
\section*{Introduction}

Learning rate schedules are used across all areas of modern machine learning, yet very little is known on which schedule is most suited for a given problem. This question has been thoroughly studied for convex problems, where the optimal learning rate schedule generally goes as $\lr(t)\tsim 1/t$~\cite{liu2019deep,li2017stochastic}. However, deep neural networks and other high-dimensional modern optimization problems are known to operate in highly non-convex loss landscapes~\cite{brea2019weight,choromanska2015loss}. Developing a theory to understand the impact of scheduling in this setting remains a crucial challenge. 

In this work we present, to the best of our knowledge, the first analytical study of this problem for gradient-based algorithms. We focus on the high-dimensional inference problem of retrieving a ground truth signal $\signal\in\R^N$ from observations via a noisy channel. When the noise dominates the signal, the loss simply boils down to a Gaussian random function on the $N$-dimensional sphere ($N\to\infty$). This optimization problem 
has been studied in the literature for constant learning rate, both using rigorous methods and techniques from statistical physics, see~\cite{arous2006cugliandolo, dembo2020dynamics,arous2020algorithmic, mannelli2020thresholds,zdeborova2016statistical}  and references therein.

\begin{figure}
    \centering
    \includegraphics[width=0.8\linewidth]{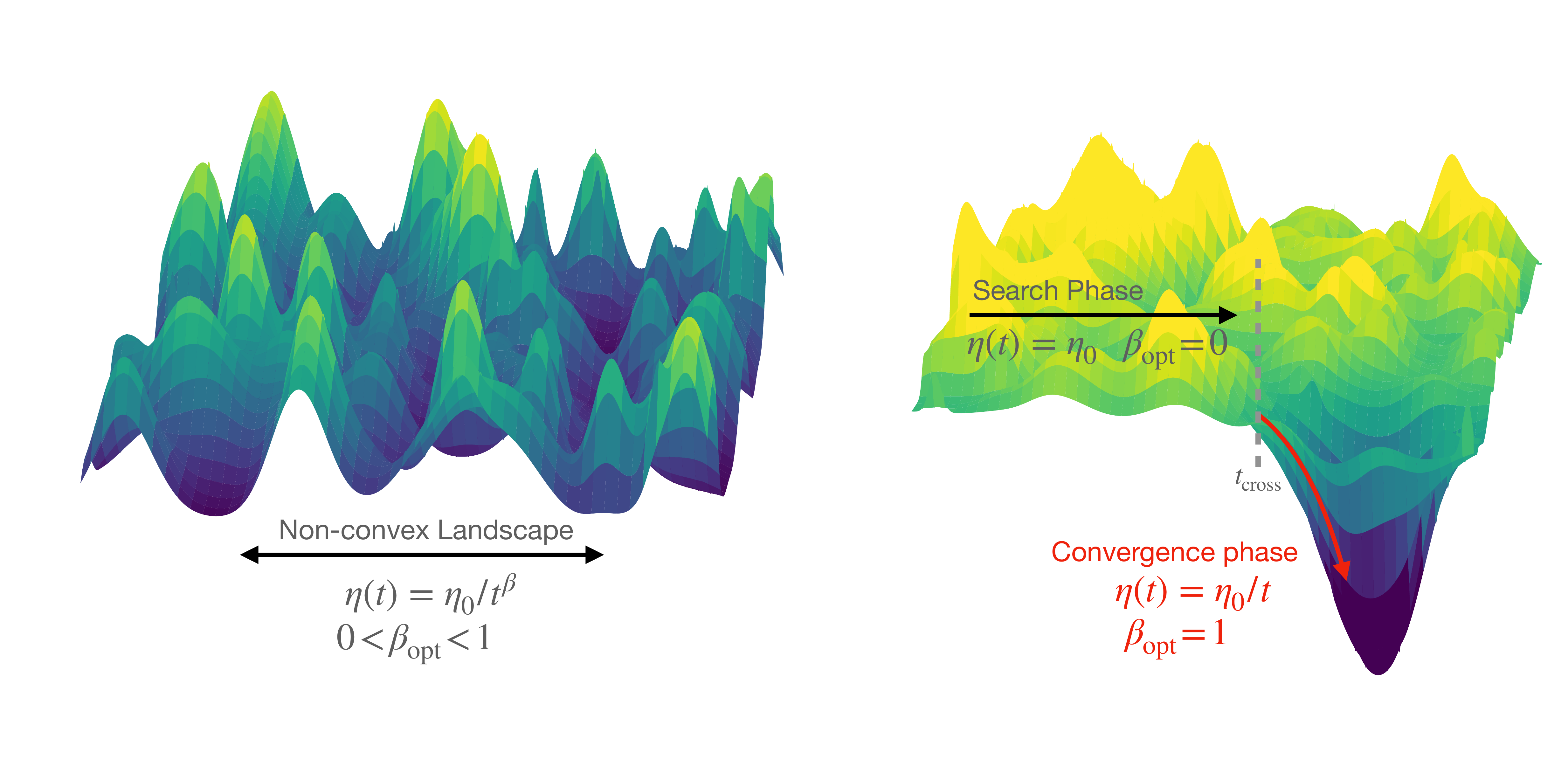}
    \caption{\textbf{The optimal learning rate schedule depends on the structure of the landscape.} \emph{(Left):} in the purely non-convex landscapes of Sec.~\ref{sec:unplanted}, the learning rate must be decayed as $\lr(t)=\lr_0/t^\lrexponent$ with $\lrexponent<1$ to speed up optimization. \emph{(Right):} the landscapes of Secs.~\ref{sec:plantedmodel} and~\ref{sec:TS} feature basins of attraction due to the presence of a signal to recover. One must first keep a large constant learning rate to escape the rough parts of the landscape as quickly as possible, then decay the learning rate as $\lr(t)=\lr_0/t$ once inside a convex basin.}
    \label{fig:sketch}
    \vspace{-.5cm}
\end{figure}

\paragraph{Setup}

Learning rate decay is generally used to reduce the noise induced by optimization schemes used in practice. For example, SGD with batch size $\bs$ typically induces a noise which scales as the learning rate divided by batch size  $\nicefrac{\lr}{\bs}$~\cite{jastrzkebski2017three,park2019effect,smith2017don,mignacco2021effective}. To mimick this optimization noise, we focus on Langevin dynamics \cite{cheng2020stochastic, mingard2021sgd, hu2017diffusion, li2017stochastic}. Given a loss function $\loss$ and a temperature $T$, this consists in minimising $\mathcal{L}$ by updating the estimate $\spin\in\mathbb R^N$ of the signal from a random initial condition according to the equation:
\begin{align}
\label{eq:langevin}
    \begin{split}
        \frac{\dd \spin_i(t)}{\dd t} = - \lr(t)\left(\frac{\partial \loss(\spin, \signal)}{\partial \spin_i} + \noise_i(t) + \spherical(t)\spin_i(t) \right),
    \end{split}
\end{align}
where $\noise(t)\in\R^N$ is a Gaussian noise with $0$ mean and variance $\langle \noise_i(t) \noise_j(t^{\prime})\rangle = 2 T\delta_{ij} \delta(t-t^{\prime})$, and the Lagrange multiplier $z(t)$ is used to enforce the spherical constraint $\Vert\spin\Vert^2=N$ which we impose throughout the paper ($z(t)$ can be thought of as a weight decay that evolves during training to keep the norm of the estimator fixed). The temperature $T$ represents the strength of the noise inherent to the optimisation algorithm, i.e. $1/\bs$ for SGD (we consider $T<1$ in the following). To study scheduling, we decay the learning rate as $\lr(t)=\lr_0/t^{\lrexponent}$, as commonly chosen in the literature~\cite{moulines2011non,xu2011towards}. Note that here we are considering gradient-flow -- our results are confirmed by experiments performed with gradient descent.

We consider two models for the loss $\loss$: the (planted) \emph{Sherrington-Kirkpatrick} (SK) model~\cite{sherrington1975solvable}, where the signal is scrambled by a random matrix, and the more involved \emph{spiked matrix-tensor} (SMT) model~\cite{mannelli2020marvels}, where the signal is additionally observed through its contraction with a random tensor of order $p$. The first setup is analytically tractable both at infinite and finite dimensions~\cite{cugliandolo1995full,barbier2021finite}, and its landscape features a number of critical points which grows linearly with the dimension. The second setup is more involved and requires a mean-field treatment in the infinite dimensional limit~\cite{cugliandolo1993analytical,mannelli2019afraid,mannelli2020marvels}. The number of critical points grows exponentially with the dimension and has been studied analytically with the \emph{Kac-Rice} method~\cite{arous2019landscape,ros2019complex}. This distinction allows us to grasp how the amount of non-convexity impacts the optimal decay of the learning rate. 

\paragraph{Contributions}

We begin by considering the purely non-convex setup where the signal is undetectable (left panel of Fig.~\ref{fig:sketch}). The loss is then a Gaussian random function on the $N$-dimensional sphere with zero mean and a covariance $
\mathbb{E}[\loss(x)\loss(x')]\propto(x\cdot x')^p$. We determine the optimal learning rate to reach the lowest value of the loss function on an arbitrarily large (but finite) time in the high-dimensional limit. For the $p=2$ case, corresponding to the spherical SK model, we find $\beta~=~1/2$ whereas for $p>2$ we obtain $\beta~=~2/5$. The higher degree of non-convexity of the latter requires the learning rate to be decayed more slowly; we generalize these findings by leveraging results from out-of-equilibrium physics. Note that inverse square root decay is commonly used among practitioners in state-of-the-art endeavours such as training Transformers~\cite{vaswani2017attention}; our analysis provides theoretical evidence for its soundness in a particular class of non-convex landscapes.

We then study the influence of a detectable signal (right panel of Fig.~\ref{fig:sketch}), and we determine the optimal learning rate schedule to find the signal in the shortest amount of time. 
In this case, a crossover time emerges between two phases~\cite{bottou2003stochastic}: a {\it search} phase, where the signal is weak and the dynamics travel through a rugged landscape, followed by a  {\it convergence} phase the signal is detected and the problem becomes locally convex. We show that the optimal schedule is to keep a large constant learning rate during the first phase to speed up the search, then, once in the convex basin, to decay the learning rate as $\nicefrac{1}{t}$. This protocol allows to speed up convergence and find lower loss solutions, and is reminiscent of schedules used in practice.

Finally, we show through experiments that these insights are reflected in practice when training neural networks on a teacher-student regression task with SGD. 

\paragraph{Related work}

Typically, learning rate schedules consist in a large learning rate phase followed by a decay phase. A body of works have shown that this allows to learn easy patterns early on and complex patterns later~\cite{you2019does,li2019towards}. Although stepwise decays of the learning rate were used for a long time~\cite{he2016deep,ge2019step}, most recent works have turned to smooth decays such as inverse square root~\cite{vaswani2017attention} and cosine annealing~\cite{loshchilov2016sgdr}, which involve less hyperparameters to tune. Other possibilities include cyclical learning rates~\cite{smith2017cyclical} and automatic schedulers~\cite{lewkowycz2021decay}. 

The use of a warmup~\cite{goyal2017accurate} before decaying the learning rate has shown to be effective in avoiding instabilities arising from large learning rates~\cite{gilmer2021loss,gotmare2018closer}.  
Another common practice is to use adaptive optimizers, which select a different learning rate for each learning parameter~\cite{kingma2014adam,zeiler2012adadelta,duchi2011adaptive}, although these have been shown to often degrade generalization~\cite{keskar2017improving,chen2018closing,wilson2017marginal}. 

On the theoretical side, several works have studied Langevin dynamics for mean-field spin glasses. Particularly relevant to us are those which focus on the spherical SK setup \cite{barbier2021finite,cugliandolo1995full}, as well as those showing the existence of a search and a convergence phase for the SMT model \cite{mannelli2020marvels}. However, to the best of our knowledge, no previous works have studied these kind of highly non-convex optimization problems in the context of a non-constant learning rate. Our analysis is based on common methods in theoretical physics which have been to a large extent made rigorous in recent years \cite{arous2006cugliandolo, arous2019landscape, dembo2020dynamics, mannelli2019afraid}, and is confirmed by numerical experiments.   

\paragraph{Reproducibility} The code to reproduce the figures in this paper is available at~\url{https://github.com/mariaref/nonconvex-lr}.

\section{The speed-noise trade-off in a simple convex problem}
\label{sec:convex}
Before studying non-convex problems, it is instructive to recall the effect learning rate decay has on optimisation in a simple 1D convex basin of curvature $\kappa$, for which $\loss(\spin)~=~\frac{1}{2} \kappa \spin^2$. 
The Langevin equation (Eq.~\ref{eq:langevin}) can easily be solved and yields (see App.~\ref{app:convex}):
\begin{align}
   \langle \loss(t) \rangle = &\underbrace{\frac{\kappa\spin(t_0)^2}{2} e^{-2\kappa \int_{t_0}^t \dd \tau \lr(\tau)}}_{\bar \loss(t)} \\
   &+ \underbrace{\frac{\kappa T}{2} \int_{t_0}^t \dd t' \lr(t')^2 e^{-2\kappa\int_{t'}^{t} \dd \tau \lr(\tau)}}_{\delta \loss(t)}, \notag
\end{align}
where $\langle . \rangle$ denotes an average over the noise $\xi$.
The first term is an \emph{optimization} term, which amounts to forgetting the initial condition $\spin(t_0)$. It is present in absence of noise ($T=0$) and its decrease is related to the way the dynamics descend in the loss landscape. The second term is a \emph{noise} term, which is proportional to the strength of the noise $T$, and reflects the impact Langevin noise has on optimization. 

To converge to the solution $\spin=0$ as quickly as possible, one is faced with a dilemma: reducing the learning rate suppresses the effect of the noise term $\delta \loss$, but also slows down the dynamics, leading to a larger optimization term $\bar \loss$. The ideal tradeoff is found when these two effects are comparable.
By taking $\lr(t)=\lr_0/t$ we obtain:
\begin{align}
    \bar \loss(t) \propto t^{- 2\lr_0\kappa}, \qquad \delta \loss(t) \propto 1/t.
    \label{eq:convex}
\end{align}
Hence, the loss decays to zero as $\nicefrac{1}{t}$ if we take $\lr_0\geq\nicefrac{1}{2\kappa}$, as found in many previous works~\cite{liu2019deep,li2017stochastic}. 
Note that if we take a slower decay such as $\lr(t)\tsim \nicefrac{1}{t^\lrexponent}$ with $\lrexponent<1$, $\bar \loss(t)$ converges to 0 exponentially fast, but $\delta \loss(t) \!\propto\! \lr(t)$ decays slower and bottlenecks the loss. 
Conversely, if we take a faster schedule, i.e. $\lrexponent>1$, 
then the noise term decays faster, but the dynamics stop before reaching the solution, as $\bar \loss(t)$ does not converge to zero when $t\to\infty$. 
 
This simple example illustrates the trade-off between the speed of optimisation and the noise suppressing effect, which will be the cornerstone of proper scheduling in the high-dimensional non-convex settings studied below.

\section{Optimal decay rates in random landscapes}
\label{sec:unplanted}

In this section, we consider purely non-convex optimization landscapes, where the loss $\loss$ is a Gaussian random function defined on the $N$-dimensional sphere, with zero mean and covariance: 
\[
\mathbb{E}[\loss(x)\loss(x')]=\frac{N}{2}(x\cdot x')^p, \quad p\geq 2.
\]
This setup, which has been studied in great detail in the context of statistical physics, can be viewed as a special case of the inference problems of Sec.~\ref{sec:plantedmodel} where the noise is too strong for the signal to be detectable. The aim is not to retrieve a signal, but simply to decrease the loss as quickly as possible on an arbitrarily large (but finite) time. 


\subsection{Sherrington-Kirkpatrick model}
\label{sec:SKmodel}
We start by focusing on the case $p=2$.
This can be achieved with the spherical version of the spin glass model introduced by~\citet{sherrington1975solvable}. Here, the variables $\spin_i$ and $x_j$ interact with each other via random symmetric couplings\footnote{As discussed in App.~\ref{app:SK}, due to the universality typical of random matrix theory distributions, our results hold for a broad class of distributions for the couplings. Note also that the diagonal terms do not matter in the large N limit but for simplicity we take $\coupling_{i i}~\tsim~\mathcal{N}(0,2)$.} $\coupling_{i j}~\tsim~\mathcal{N}(0,1)$, and, as throughout the paper, are required to satisfy the spherical constraint $\Vert\spin(t)\Vert^2=N$. The loss function is given by:
\begin{align}
    \begin{split}
    \label{eq:loss_SK}
        \loss(\spin) = -\frac{1}{\sqrt{N}} \sum_{i<j}^N J_{i j} \spin_{i} \spin_{j}. 
    \end{split}
\end{align}

In this section, we consider the high-dimensional limit $N\to\infty$; finite-dimensional effects are discussed in Sec.~\ref{sec:plantedSKmodel}.

\paragraph{Solving the dynamics}

To obtain the value of the loss function at all times, we multiply the original Langevin equation by $\spin_i$ and sum over all components. Using Ito's lemma, and the concentration of $z(t)$ in the $N\to\infty$ limit, leads to the simple relation:
\begin{align}
\label{eq:loss_SK_noise}
    0 = \left\langle\frac{\partial \Vert \spin \Vert^2}{\partial t}\right\rangle
    &= \lr(t) \left[ - 2 \loss(t) - N \spherical(t) \right] + N \lr(t)^2 T\notag\\
    \Rightarrow \loss(t) &= -\frac{N}{2} \left(\spherical(t) - \lr(t) T\right)
\end{align}

As in the convex setup, we find a competition between an optimization term and a noise term. Since the temperature is fixed, the latter decays as $\lr(t)$. To obtain the value of the Lagrange multiplier $z(t)$, we impose the spherical constraint at all times, yielding (see App.~\ref{app:SK}):
\begin{align}
    \spherical(t) &= 2 - \frac{3(1-\lrexponent)}{4t^{1-\lrexponent}}.
\end{align}

Hence, the scaled loss $\ell = \nicefrac{\loss}{N}$ converges to the ground state (global minimum) $\ell_{GS}\!~=~\!-1$ as a sum of power-laws:
\begin{align}
    \ell(t) - \ell_{GS} &= \frac{\lr_0 T}{2t^\lrexponent} + \begin{cases}
    \frac{3(1-\lrexponent)}{8 \lr_0 t^{1-\lrexponent}}, \quad \beta<1\\
    \frac{3}{8 \lr_0 \log t}, \quad \beta=1
    \end{cases}.
    \label{eq:decay_SK_model}
\end{align}


\begin{figure}[tb]
    \centering
    \includegraphics[width=0.8\linewidth]{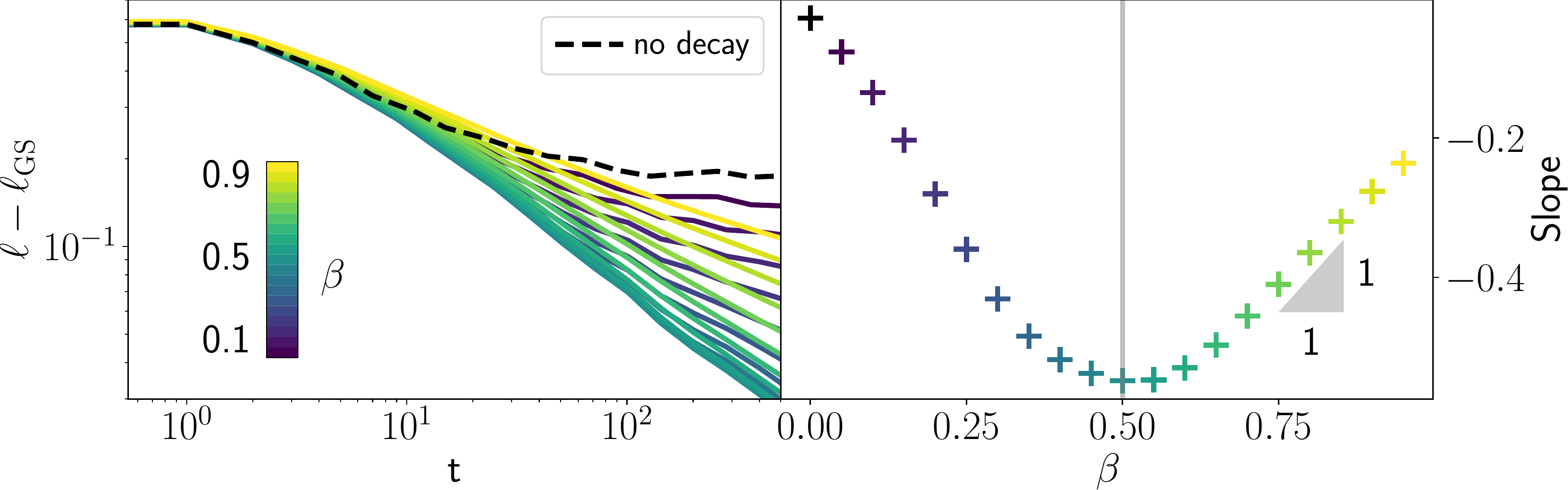}
    \caption{\textbf{In the SK model, the optimal decay rate is $\beta_{opt}=0.5$.} (\emph{Left}) Loss curves of the SK model when decaying the learning rate as $\lr(t)=\lr_0/t^\lrexponent$ for various values of $\lrexponent$ (colored lines). (\emph{Right}) Decay exponent of $\ell-\ell_\mathrm{GS}$ at long times as a function of $\lrexponent$. We recognize a decay exponent of $\min(\lrexponent, 1-\lrexponent)$ as predicted by Eq.~\ref{eq:decay_SK_model}, which is fastest for $\beta_\opt=\nicefrac{1}{2}$. \emph{Parameters:} $N=3000$, $T=1$, $\lr_0=0.1$.}
    \label{fig:exponents}
\end{figure}

\paragraph{Optimal decay rate}
At long times, Eq.~\ref{eq:decay_SK_model} implies a power-law decay of the loss with an exponent $\min(\lrexponent, 1-\lrexponent)$ due to the speed-noise tradeoff. Hence, the optimal decay rate at long times is $\beta_\opt\!~=~\!\nicefrac{1}{2}$. This is confirmed by numerical simulations at finite size, see Fig.~\ref{fig:exponents}. Note that this decay rate is empirically chosen to train many state-of-the-art neural networks such as the original Transformer~\cite{vaswani2017attention}, but, to the best of our knowledge, has never been justified from a theoretical point-of-view in a non-convex high-dimensional setting.

\paragraph{Curvature analysis}
To gain better understanding, it is informative to study the local curvature of the effective landscape the dynamics take place in. To do so, one needs to compute the spectrum of the effective Hessian taking into account the spherical constraint of Eq.~\ref{eq:langevin}, defined as:
\begin{align}
    \hessian\!~=~\!\frac{1}{\sqrt N}\coupling+\spherical(t) I.
    \label{eq:hessian}
\end{align} 
In the the $N\to\infty$ limit, the spectral density of the first term, defined as $\rho(\mu)\!~=~\!\sum_{i=1}^N\!~ \delta(\mu-\mu_i)$, converges to a semi-circle law~\cite{wigner1958distribution}:
\begin{align}\label{eq:sc}
    \rho_{sc}(\mu)= \frac{1}{2 \pi} \sqrt{4-\mu^{2}}, \quad \forall \mu\in[-2, 2].
\end{align} 

\begin{figure}
    \centering
    \includegraphics[width=0.8\columnwidth]{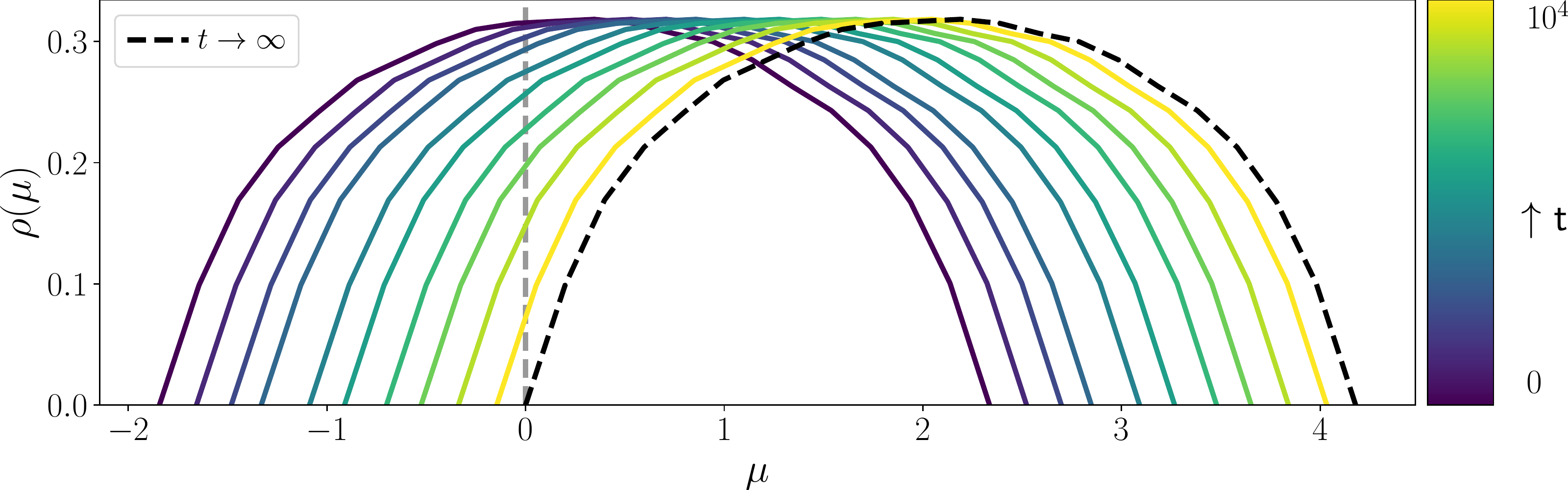}
    \caption{\textbf{In the SK model, the dynamics never reach a convex region.} During training, the local curvature, i.e. the spectral density of the Hessian (Eq.~\ref{eq:hessian}) shifts to the right. The left hand side of the spectrum only reaches 0 at $t\to\infty$, signalling that there remains negative eigenvalues at any finite time. \emph{Parameters:} $N=3000$, $T=1$, $\eta_0=0.1$, $\beta=0.8$. }
    \label{fig:hessian}
\end{figure}

The spectral density of $\hessian$ is shifted to the right during the dynamics by the Lagrange multiplier $z(t)$, reflecting the way in which the local curvature changes with $t$. 
As show in Fig~\ref{fig:hessian} and known from previous works~\cite{cugliandolo1995full}, there remains negative eigenvalues at any finite time: the right edge of the spectrum only reaches 0 asymptotically as $t\to\infty$, since $z(t)\to2$.  

Hence, the dynamics never completely escape the saddles of the landscape at $N\rightarrow \infty$. This ruggedness of the landscape entails slow ``glassy" dynamics, characterized by a power-law decay of the optimization term for any $\beta<1$, contrary to the exponential decay obtained in the convex setup (Sec.~\ref{sec:convex}). 

\subsection{The $p$-spin model}
\label{sec:pspinmodel}

We now turn to the analysis of the $p$-spin model which has been extensively studied in physics as a model of structural glasses, see e.g.~\cite{berthier2011theoretical}. To us, it is an ideal candidate as it corresponds to a random Gaussian landscape (with $p>2$) for which the \emph{Kac-Rice} approach rigorously shows the existence of a number of critical points growing exponentially with the dimension~\cite{auffinger2013random}. It is thus intrinsically harder, i.e. more strongly non-convex than the SK model above. The loss of the $p$-spin model (for $p>2$) is written as:
\begin{equation}
    \label{eq:Hpdspin}
    \loss = -\sqrt{ \frac{(p-1)!}{N^{p-1}}} \sum_{i_1<...<i_p} \coupling_{i_1...i_p} \spin_{i_1}... \spin_{i_p}. 
\end{equation}

\paragraph{Solving the dynamics}
In the high-dimensional limit $N\to\infty$, the Langevin dynamics of the system can be reduced to a closed set of PDEs for a set of ``macroscopic" quantities, which concentrate with respect to the randomness in the couplings $J$ and the thermal noise in the dynamics $\noise$, as shown rigorously in \cite{arous2006cugliandolo}.
These quantities are the two-point correlation $C(t,t')$ of the system at times $t,t'$ and the response $R(t,t')$ of the system at time $t$ to a perturbation in the loss function at an earlier time $t'$:
\begin{align}
    C(t,t') &= \lim_{N\to\infty}\frac{1}{N}\mathop{\EE}_{\noise, \coupling}\sum_{i=1}^N \spin_i(t)\spin_i(t'),\\
    R(t,t') &= \lim_{N\to\infty}\frac{1}{N}\mathop{\EE}_{\noise, \coupling}\sum_{i=1}^N\frac{\delta \spin_i(t)}{\delta \noise_i(t')}.
\end{align}
Their dynamics is described by a closed set of integro-differential equations, dubbed the Crisanti-Horner-Sommers-Cugliandolo-Kurchan (CHSCK) equations \cite{cugliandolo1993analytical, crisanti1992spherical,arous2006cugliandolo}. We extend these equations to the non-constant learning rate case using the methods reviewed in \cite{castellani2005spin}: 
\begin{align}
\label{eq:dynamical_equations_no_signal}
&\frac{\partial R\left(t_{1}, t_{2}\right)}{\partial t_{1}}= F^p_R(\spherical, R, C, \lr),
\\
&\frac{\partial C\left(t_{1}, t_{2}\right)}{\partial t_{1}}= F^p_C(\spherical, R, C, \lr),\\
&\spherical(t)\!=\! T\lr(t) + p\int dt_2 \lr(t_2) R(t_2, t) C^{p-1}(t_2, t),
\end{align}
where we deferred the full expression of the update functions $F^p_R$ and $F^p_C$ as well as their derivation to App.~\ref{app:derivationPDE}. 

Imposing the the spherical constraint $C(t,t)\!~=\!~1$ allows to find the value of the spherical constraint $z(t)$. To compute the loss, we follow the same procedure as in the SK model and obtain:
\begin{align}
\label{eq:loss_pspin_no}
   \ell(t)\equiv \frac{\loss}{N} = -\frac{1}{p}\left(  \spherical(t) - T \lr(t) \right ).
\end{align}

\paragraph{Optimal decay rate}
Here again we find that two competing terms contribute to the loss, the first related to optimisation and the second to noise. By choosing a learning rate $\lr(t)=\nicefrac{\lr_0}{t^{\lrexponent}}$, the later decays as $t^{-\lrexponent}$. 
The decay of the former is more complex due to the high complexity of the landscape. It can be shown \cite{cugliandolo1993analytical} that the system never reaches the ground state, instead remaining trapped in so-called threshold states where the Hessian has many zero eigenvalues (the density of eigenvalues is a Wigner semicircle whose left edge is zero as in the SK model). The loss is then given by:
\begin{equation}
   \ell_\th= -\frac{\sqrt{4(p-1)}}{p} > \ell_{GS}.
   \label{eq:loss_threshold}
\end{equation}

The relaxation towards the threshold states is characterised by a power-law due to the rough energy landscape, but with a different exponent this time: $\spherical_\th - \spherical(t)~\propto~t^{-\gamma}$, with $\gamma=\nicefrac{2}{3}$ at $T=0$~\cite{thalmann2001geometrical}. 
Using the CHSCK equations (\ref{eq:dynamical_equations_no_signal}), we analytically show in App.~\ref{app:pspin} that with decaying learning rate the exponent becomes $\gamma(1-\beta)$.
Hence, similarly to the SK model, the decay of the loss is controlled by a competition between two power-laws:
\begin{align}
    \ell(t) - \ell_\mathrm{th} &\sim t^{-\min(\beta, \agingexp(1-\beta))}\notag\\
    \Rightarrow\lrexponent_\opt &= \frac{\gamma}{1+\gamma} = \frac{2}{5}.
    \label{eq:betaopt_sspin}
\end{align}


\begin{figure}[tb]
    \centering
        \includegraphics[width=0.8\linewidth]{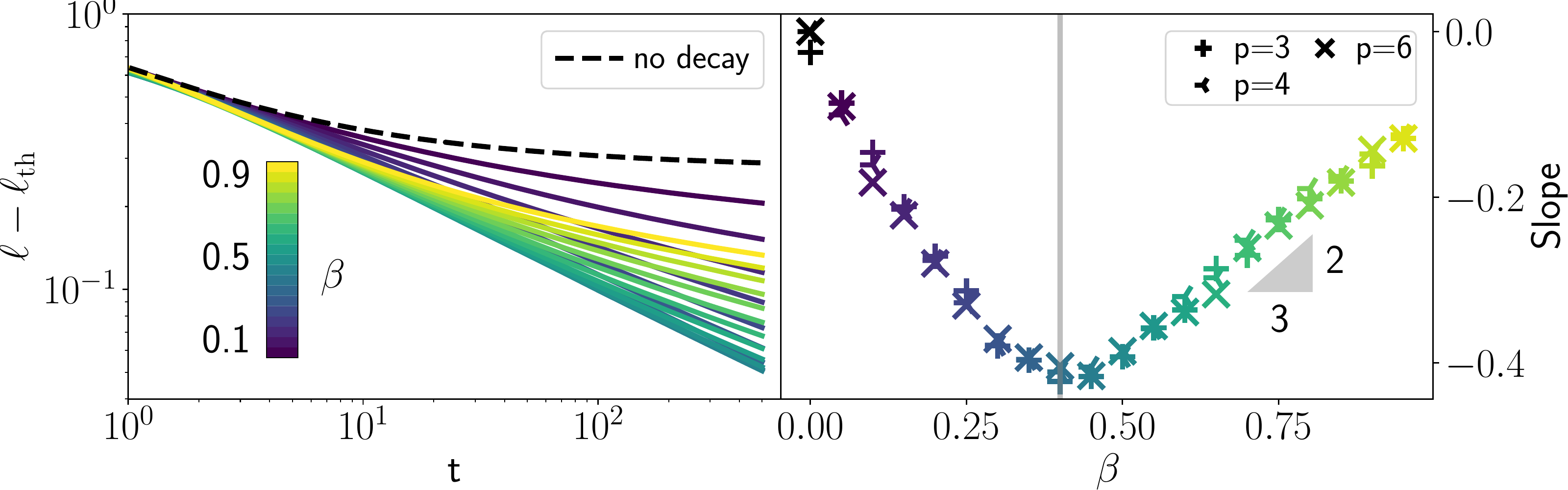}  
        \caption{\textbf{In the $p$-spin model, the optimal decay rate is $\lrexponent_{opt}~=~0.4$}. \emph{(Left)} Loss curves of the 3-spin model at $T=1$ when decaying the learning rate as $\lr(t)=\lr_0/t^\lrexponent$ for various values of $\lrexponent$ (colored lines).  \emph{(Right)} Decay exponent of $\ell-\ell_\th$ at long times, for various $p$, as a function of $\lrexponent$. We recognize a decay exponent of $\min(\beta, \gamma(1-\beta))$, as predicted by Eq.~\ref{eq:betaopt_sspin}, which is fastest for $\beta_\opt=\nicefrac{2}{5}$. 
        \emph{Parameters: }$dt=10^{-2}$, $\lr_0=0.5$, $T=1$. }
    \label{fig:pspin_noSignal}
\end{figure}

In Fig.~\ref{fig:pspin_noSignal}, we numerically integrate Eqs.~\ref{eq:dynamical_equations_no_signal} for $p=3, 4, 6$, confirming that the optimal decay rate to balance the noise and the optimization terms is $\beta_{opt}=\nicefrac{2}{5}$. The numerical integration is non-trivial and we implement it using the tools developed in \cite{mannelli2020marvels}.

\subsection{Relation with annealing in physics}
The results found in this section can be put in a very general framework that was developed in physics of out of equilibrium systems. As shown in App.~\ref{app:general_scaling}, using a learning rate schedule is equivalent to annealing the temperature of the physical system as a power-law $t^{\nicefrac{-\lrexponent}{1-\lrexponent}}$. Thus, finding the optimal learning rate schedule to minimize the loss is equivalent to determining the optimal annealing protocol to decrease the energy. 
A key ingredient in the solution is how fast the dynamics descend in the loss landscape in absence of noise. In physical systems, this optimization term generally follows a power-law decay with exponent $\gamma$~\cite{bray2002theory,bouchaud1998out,biroli2005crash}. 

At finite temperature, the speed-noise tradeoff requires this decay rate to be equal to that of the temperature, $\nicefrac{\lrexponent}{1-\lrexponent}$, leading to $\beta_\mathrm{opt}=\nicefrac{\gamma}{1+\gamma}$. The exponent $\gamma$ has been determined in many statistical physics problems, corresponding to different high-dimensional non-convex landscapes, and typically ranges from zero (logarithmic relaxation) to one. Our results extend to all these problems and, and predict optimal annealing exponents varying between $0$ and $\nicefrac{1}{2}$.

\section{Recovering a signal: the two phases of learning}
\label{sec:plantedmodel}

We now move to the setup where there is a signal $\signal$ in the problem, which the algorithm aims to retrieve in the shortest time possible. In addition to the random Gaussian function, the loss now contains a deterministic term forming an attraction basin in the landscape, as sketched in the right panel of Fig.~\ref{fig:sketch}. 

\subsection{Spiked Sherrington-Kirkpatrick model}
\label{sec:plantedSKmodel}

We first consider the so-called planted SK model, where the objective is to retrieve a ground truth $\signal$ such that $\Vert\signal\Vert^2=N$, i.e. maximize the overlap with the signal $m= \nicefrac{\sum_i\spin_i\cdot\signal_i}{N}$. We enforce as before the spherical constraint $\Vert\spin\Vert^2=N$ which induces $m\in[-1,1]$, and sample randomly the initial configuration of $\spin$, such that the initial overlap is of order $\nicefrac{1}{\sqrt N}$. The loss function takes the form:
\begin{align}
    \begin{split}
    \label{eq:loss_planted_SK}
        \loss(\spin) &= - \frac{N}{2} m^2-\frac{\Delta}{\sqrt{N}} \sum_{i<j}^N \coupling_{i j} \spin_{i} \spin_{j} = \frac{1}{2}\spin H \spin^\top,
    \end{split}
\end{align}

with $H\!=\!- \frac{\Delta}{\sqrt N} J - \frac{1}{N}\signal{\signal}^\top$.

Decreasing $\Delta$ makes the signal easier to detect, leading to an easier problem. For $\Delta\!~<~\!\nicefrac{1}{2}$, an eigenvalue of $H$ pops out of the semicircle law (\ref{eq:sc}) as a BBP transition takes place \cite{baik2005phase}, leading to the follow spectrum:
\begin{align}
    \rho(\mu) = \left(1-\frac{1}{N}\right) \rho_{sc}(\mu/\Delta) + \frac{1}{N}\rho(\mu-1)
\end{align}
This is the regime in which the signal overcomes the noise, i.e. the global minimum of the loss has a finite overlap with the signal, which can then be retrieved by gradient flow (or gradient descent).

In the following, we assume that $\Delta<\nicefrac{1}{2}$, and define the gap between the largest and second largest eigenvalue as $\kappa \equiv 1-2\Delta$. In App.~\ref{app:SK}, we analytically show the emergence of a crossover time,
\begin{align}
  \tjump = 
    \left( \frac{\log N}{2\lr_0 \kappa }\right)^{\frac{1}{1-\lrexponent}}.
    \label{eq:tjump}
\end{align}
Before $\tjump$, the system behaves as if the signal was absent, i.e. as in Sec.~\ref{sec:SKmodel}: this is the \emph{search} phase. After $\tjump$, the signal is detected: this is the \emph{convergence} phase. The loss becomes: \begin{align}
    \ell(t)-\ell_{GS} &= \frac{\lr_0 T}{2t^{\lrexponent}} + \begin{cases}
        \mathcal{O}(e^{-2\lr_0\kappa t^{1-\lrexponent}}), \quad &\beta<1\\
        \mathcal{O}(t^{-2\lr_0\kappa}), &\beta=1\\
    \end{cases}
\end{align}
with $\ell_{GS}=-1$. We recognize here the exact same result as obtained in the convex setup of Eq.~\ref{eq:convex}: as long as $\lr_0~>~\nicefrac{1}{2\kappa}$, the optimal learning rate schedule is $\lr~=~\lr_0/t$. This indicates that the dynamics has entered a convex basin of curvature $\kappa$. 

\paragraph{Optimal learning rate schedule}

\begin{figure*}[htb]
    \centering
    \includegraphics[width=\linewidth]{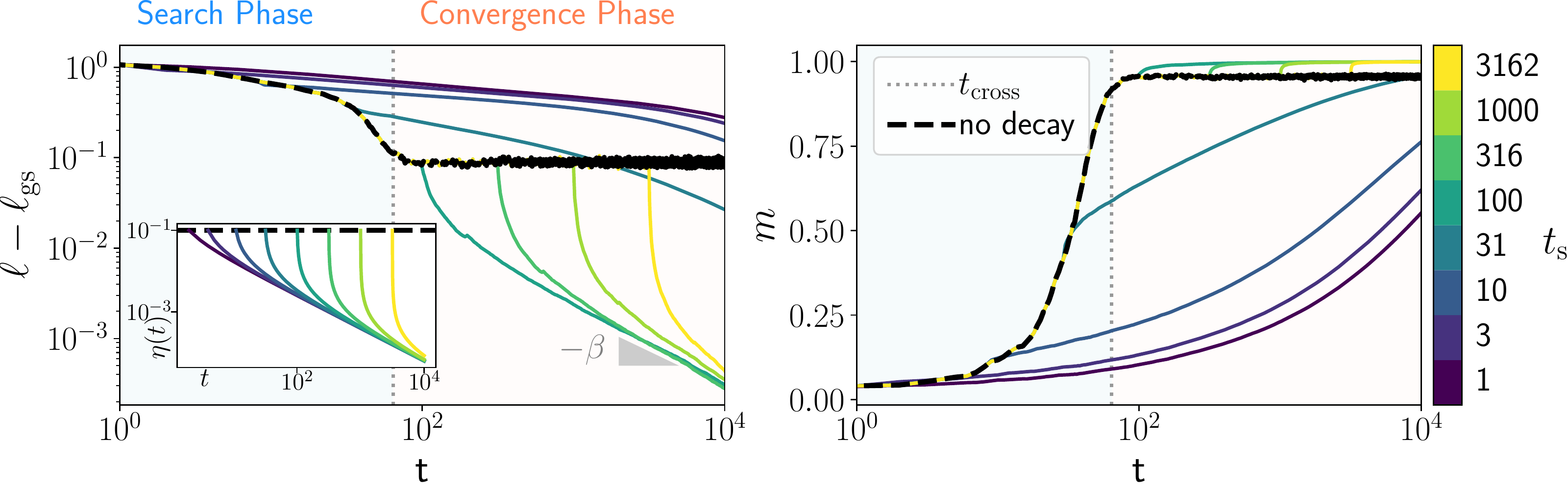}
    \caption{\textbf{Emergence of a crossover time in the planted SK model.} The dashed black show the loss \textbf{(left)} and overlap with the signal \textbf{(right)} at constant learning rate $\lr_0=0.1$. 
    The colored lines, show the result of keeping the learning rate constant until $t_s$ and then decaying as $\lr(t)=\lr_0/(t-t_s)^\lrexponent$ (as shown in the inset). The dashed vertical line marks the theoretical crossover time $\tjump=\frac{\log N}{2 \lr_0 \kappa}$, it matches with the time at which the loss, at constant learning rate, saturates. Before the crossover, decaying the learning rate is detrimental. After the crossover, it allows the model to converge to zero loss as $t^{-\lrexponent}$ and to perfectly recover the signal.
    We set $N=3000, T=1., \lr_0=0.1, \lrexponent=0.8, \kappa=0.5$.}
    \label{fig:crossover}
\end{figure*}

To speed up the initial phase where the signal hasn't yet aligned with the signal, one needs to reduce $\tjump$, which is achieved by using a large learning rate $\lr_0$ without any decay ($\lrexponent=0$). Passed this crossover, the system enters a convex basin, and the optimal exponent becomes $\lrexponent=1$. Ergo, the best schedule is to keep the learning rate constant up to $\tjump$, then to decay it with $\beta=1$, in contrast with the case without signal where $\lrexponent=\nicefrac{1}{2}$ was optimal, see Sec.~\ref{sec:SKmodel}. This is confirmed by the numerical experiments of Fig.~\ref{fig:crossover}, where we start decaying the learning rate as $\nicefrac{\lr_0}{(t-t_s)^{-\lrexponent}}$ for different "switch" times $t_s$. Decaying too early, with $t_s<\tjump$, slows down the dynamics, whereas $t_s>\tjump$ enables the system to reach the ground state at a rate $t^{-\lrexponent}$.

\paragraph{Finite-dimensional effects}
The two phases in the dynamics are a general feature when there is a finite gap $\kappa$ between the largest and second largest eigenvalue of $H$. In the $N\to\infty$ limit, this only occurs when $\Delta < \nicefrac{1}{2}$. However, when $\Delta>\nicefrac{1}{2}$, there is a finite gap at finite $N$ due to the discrete nature of the spectrum, which scales as $\kappa \tsim N^{-2/3}$~\cite{tracy1996orthogonal}. This induces a crossover time $\tjump\tsim N^{2/3}$. Hence, decaying the learning rate as $\lrexponent_\opt$ remains optimal for any finite time budget $t<\tjump$, but for a large budget $t>\tjump$, using the two-step schedule described in this section becomes optimal.

\subsection{Spiked Matrix-Tensor model}
\label{sec:spiked_tensor_model}

We finally move to the analysis of the SMT model for which the loss function is \cite{mannelli2019afraid}:
\begin{align}
    \label{eq:loss_p_spin_general}
        \loss(\spin) =& - \frac{N}{2 \Delta_2} m^2  
                        - \sqrt{\frac{1}{\Delta_2 N}} \sum_{i<j}     \coupling_{i,j}\spin_{i} \spin_{j} \\
                      & - \frac{N}{p \Delta_p} m^p 
                        - \sqrt{\frac{(p-1)!}{\Delta_p N^{p-1}}}\sum_{i_{1}<...<i_{p}}
                         \coupling_{i_{1},...,i_{p}} \spin_{i_{1}}..\spin_{i_{p}},\notag
\end{align}

where both $\coupling_{i j}$ and $\coupling_{i_{1},..,i_{p}}$ sampled i.i.d. from $\mathcal{N}(0,1)$. 
As understood from the loss function, the signal is observed through its contraction with a matrix and a tensor of order $p$. This model is a natural next step for our analysis: its loss landscape is extremely non-convex, but its dynamics are exactly solvable in the $N\to\infty$ limit. They can be described by a closed set of PDEs describing the dynamical evolution of the quantities $m(t)$,  $C(t,t')$, $R(t,t')$ and $\spherical(t)$ described in Sec.~\ref{sec:pspinmodel}. The derivation of these equations is deferred to the appendix \ref{app:derivationPDE}.



\begin{figure*}[htb]
    \centering
        \includegraphics[width=\linewidth]{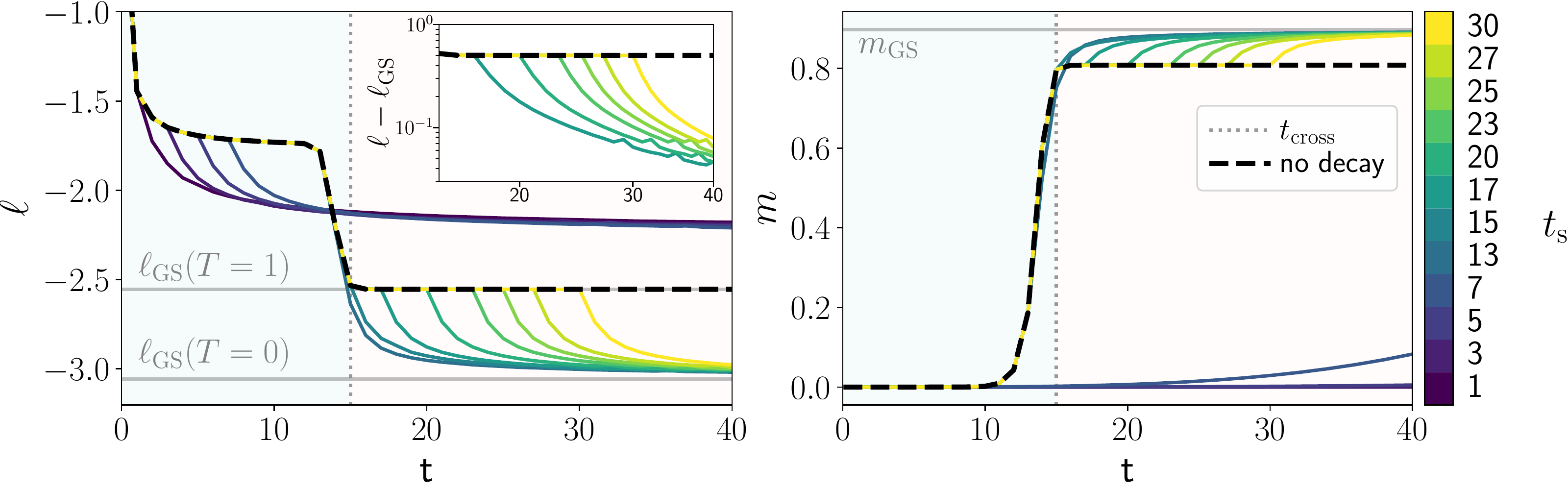}        
        \caption{\textbf{Emergence of a crossover time in the SMT model.} By fixing $\lrexponent$ from start, or anytime before $\tjump$, a randomly initialised system will remain stuck at \emph{threshold} states at high loss until $\tjump$ which is minimal for constant learning rate $\lrexponent=0$. In contrast, by decaying the learning rate at long times allows to reach lower loss solutions ({\it left}) with higher overlap with the signal ({\it right}). The optimal schedule is to keep $\lr$ constant until $\tjump$ and then set $\lrexponent=1$. By doing so, we get the best of both worlds: the first phase minimises $\tjump$ while the second allows to reach more informative solutions. {\it Parameter: $\lrexponent=0.8$, $\Delta_2=0.2$, $\Delta_p=6$, $\lr_0=1$, $T=1$, $dt = 10^{-2}$, $m_0=10^{-10}$.} }
    \label{fig:pspin_Signal}
\end{figure*}

The difficulty of the problem is controlled by the values of $\Delta_2$ and $\Delta_p$. Here, we focus on the \emph{Langevin easy} phase, defined in~\cite{mannelli2019afraid}, where a randomly initialized system recovers the signal and the overlap converges to a value of order one.\footnote{We must start from a very small initial overlap $m_0=10^{-10}$ as explained in \cite{mannelli2020marvels}, since $m_0=0$ would cause the system to remain stuck in the $N\to\infty$ limit considered here~\cite{arous2020classification}.} The dynamics in this setting have been well understood at constant learning rate in \cite{mannelli2020marvels}, and are shown as a black line in Fig.~\ref{fig:pspin_Signal} for $\lr_0=1$: the system remains trapped in the exponentially many \emph{threshold states} until a time $\tjump$. 
At $\tjump$, the system finally detects the signal and the overlap jumps to a value $m_\gs$ of order one. This behavior is reminiscent of the \emph{grokking} phenomenon observed for neural networks~\cite{power2021grokking}.

The colored lines of Fig.~\ref{fig:pspin_Signal} show that decaying the learning rate from a time $t_s$ affects optimisation in two different ways. 
(i) If we choose $t_s<\tjump$, the loss actually starts by dropping, in contrast with what was observed in Fig.~\ref{fig:crossover}. However, this drop in the loss does not yield an increase of the overlap with the signal, and the system rapidly gets stuck, remaining in a state of low overlap even after $\tjump$. 
(ii) If we choose $t_s>\tjump$, once the signal is detected, the noise is suppressed, allowing the system to converge to the ground state and the overlap to increase. Hence, the optimal schedule is again to keep a constant large learning rate during the search phase (i.e. until $\tjump$) then decay with $\beta=1$. We provide further theoretical justification for this behavior in App.~\ref{app:ground_state_loss}. 

\section{Turning to SGD : teacher-student regression}
\label{sec:TS}

Our work has demonstrated the emergence of a crossover time in a class of inference problems, before which one should keep the learning rate constant and after which it becomes useful to decay the learning rate. 

We now investigate these findings in a setup that is more realistic but simple enough to be amenable to analytical treatment in the near future.
We consider a \emph{teacher-student} regression problem in which a student network is trained to mimick the ouputs of a teacher by minimising the mean-squared error ($\mse$) over a dataset of $P$ input-outputs observations $\{\boldsymbol{x}_{\mu}, y_\mu\}\!~\in\!~\{\R^{N}, \R\}$. Here both the student $S$ and the teacher $T$ are two-layer networks: 
\begin{align*}    
    S(\spin) = \sum_{k=1}^K v_k g \left(\frac{w_m \cdot \spin}{\sqrt{N}}\right)\quad 
    T(\spin) = \sum_{m=1}^M v^\star_m g\left(\frac{w^\star_m \cdot \spin}{\sqrt{N}}\right).
\end{align*}
We train on i.i.d. gaussian inputs $x_i\tsim \normal(0,1)$ via SGD, by minimising the $\mse$ over mini-batches of size $\bs$:
\begin{equation}
    \mse = \frac{1}{\bs}\sum_{\mu=1}^{\bs}\left(S(x_\mu) - T(x_\mu)\right)^2,
\end{equation}
The optimisation noise is controlled by the batch size $B$ and is absent for full batch SGD. To study the effect of learning rate scheduling, we focus on a mini-batch of size $1$ for which optimisation noise is high. 

\begin{figure}[ht!]
    \centering
    \includegraphics[width=0.6\linewidth]{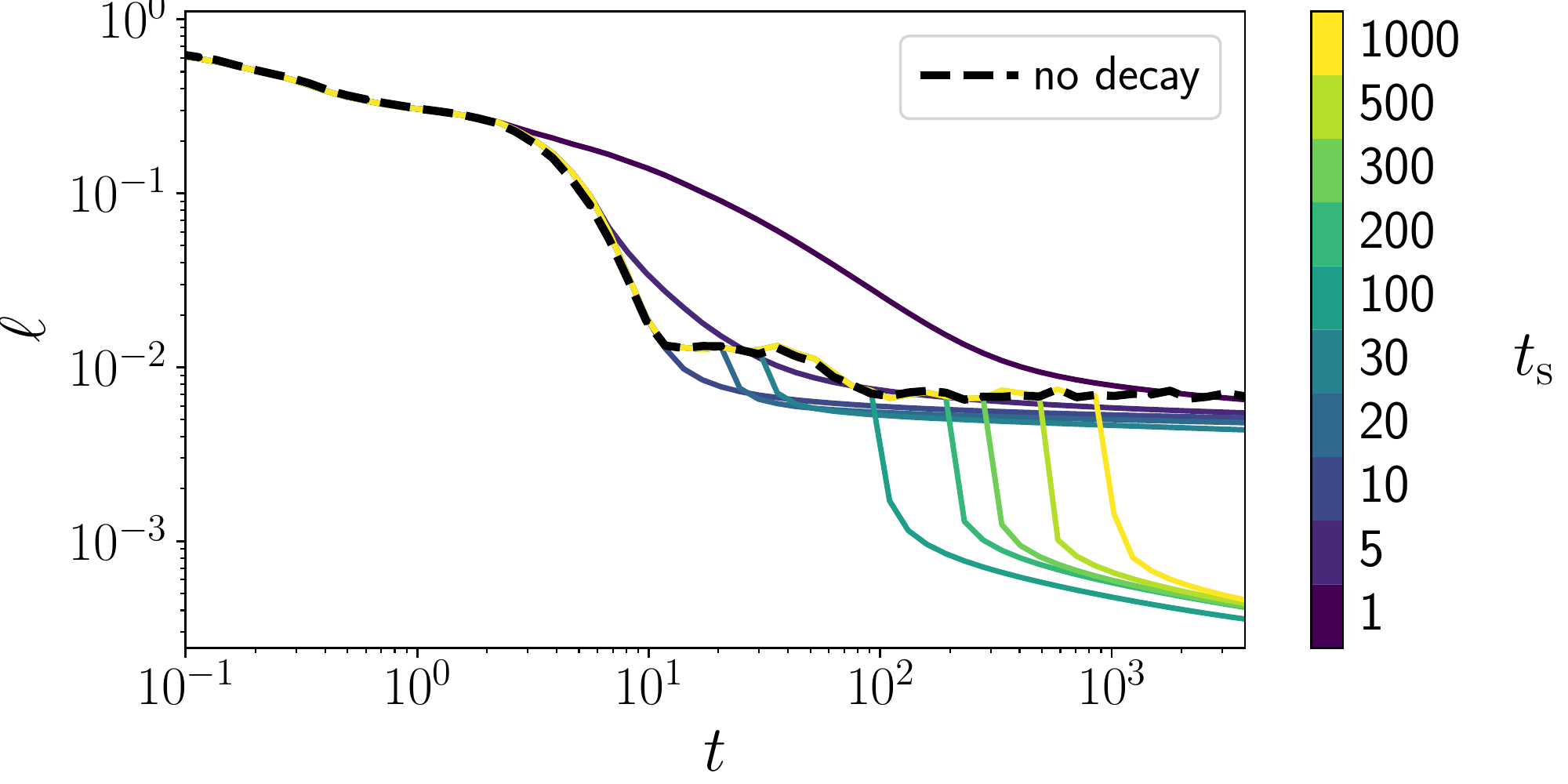}
    \caption{\textbf{The crossover time is also reflected in a regression task with SGD.} A $K$ hidden nodes 2 layer neural network student is trained to reproduce the output of her $M$ hidden nodes teacher on gaussian inputs in $N$ dimensions. As before, we find that decaying the learning rate before the loss plateaus performance, but decaying as $\lr(t)\sim t^{-1}$ once the plateau is reached allows to reach zero loss. {\it Parameters: $N=500$, $P=10^4$, $\lr_0=10^{-1}$, $M=K=2$, $\lrexponent=0.8$.}}
    \label{fig:teacher_student}
\end{figure}

Fig.~\ref{fig:teacher_student} shows the $\mse$ (calculated over the whole training set) of a student with $K=2$ hidden units learning from a teacher with $M=2$ hidden units (results with different sizes are presented in App.~\ref{app:teacher-student}). As before, we keep the learning rate constant $\lr_0$ until a time $t_s$ then decay it as $\nicefrac{\lr_0}{(t-t_s)^{-\lrexponent}}$. The phenomenology is remarkably similar to that of Sec.~\ref{sec:plantedmodel}: there exists a cross-over time $\tjump$ such that if the learning rate is decayed before $\tjump$, optimisation remains stuck at high $\mse$. In contrast, decaying the learning rate after after $\tjump$ enables to tame the noise associated with optimisation and converge to lower loss solutions. 

\section*{Conclusion}

In this work, we have analyzed learning scheduling in a variety of high-dimensional non-convex optimization problems. First, we focused on purely non-convex problems (without any basins of attraction), and showed that the optimal learning rate decay in the high-dimensional limit has an exponent smaller than one, which varies according to the degree of non-convexity of the problem at hand (ranging from $0.4$ to $0.5$ in the problems considered here). Then, we studied models where a signal must be recovered in presence of noise. In this case, what is important is not \emph{how fast} we decay the learning rate, but \emph{when} we start decaying it. 
It is better to keep a large learning rate in the {\it search} phase to find the convex basin as quickly as possible, and only then start decaying the learning rate. 

These theoretical findings are remarkably reminiscent of learning rate schedules used in practice. Establishing a tighter connection is an important direction for future work: could the $1/\sqrt t$ decay commonly used to train transformers reflect the properties of the landscape the dynamics take place in? Conversely, could one predict the optimal decay rate by inspecting the properties of the landscape? Establishing such connections in simple settings such as that of Sec.~\ref{sec:TS} is certainly within reach thanks to the recent analytical tools developed in \cite{agoritsas2018out,mignacco2020dynamical,celentano2021high,goldt2020dynamics,refinetti2021align}.
\section*{Acknowledgements}
We thank Stefano Sarao Mannelli and Francis Bach for illuminating discussions. The authors acknowledge funding from the French Agence Nationale de la Recherche under grant ANR-19P3IA-0001 PRAIRIE. 
\printbibliography

@inproceedings{vaswani2017attention,
  title={Attention is all you need},
  author={Vaswani, Ashish and Shazeer, Noam and Parmar, Niki and Uszkoreit, Jakob and Jones, Llion and Gomez, Aidan N and Kaiser, {\L}ukasz and Polosukhin, Illia},
  booktitle={Advances in neural information processing systems},
  pages={5998--6008},
  year={2017}
}

@article{tracy1996orthogonal,
  title={On orthogonal and symplectic matrix ensembles},
  author={Tracy, Craig A and Widom, Harold},
  journal={Communications in Mathematical Physics},
  volume={177},
  number={3},
  pages={727--754},
  year={1996},
  publisher={Springer}
}

@article{berthier2011theoretical,
  title={Theoretical perspective on the glass transition and amorphous materials},
  author={Berthier, Ludovic and Biroli, Giulio},
  journal={Reviews of modern physics},
  volume={83},
  number={2},
  pages={587},
  year={2011},
  publisher={APS}
}

@article{arous2006cugliandolo,
  title={Cugliandolo-Kurchan equations for dynamics of spin-glasses},
  author={Ben Arous, Gerard and Dembo, Amir and Guionnet, Alice},
  journal={Probability theory and related fields},
  volume={136},
  number={4},
  pages={619--660},
  year={2006},
  publisher={Springer}
}

@inproceedings{he2016deep,
  title={Deep residual learning for image recognition},
  author={He, Kaiming and Zhang, Xiangyu and Ren, Shaoqing and Sun, Jian},
  booktitle={Proceedings of the IEEE conference on computer vision and pattern recognition},
  pages={770--778},
  year={2016}
}

@article{celentano2021high,
  title={The high-dimensional asymptotics of first order methods with random data},
  author={Celentano, Michael and Cheng, Chen and Montanari, Andrea},
  journal={arXiv preprint arXiv:2112.07572},
  year={2021}
}

@article{mignacco2020dynamical,
  title={Dynamical mean-field theory for stochastic gradient descent in Gaussian mixture classification},
  author={Mignacco, Francesca and Krzakala, Florent and Urbani, Pierfrancesco and Zdeborov{\'a}, Lenka},
  journal={arXiv preprint arXiv:2006.06098},
  year={2020}
}

@article{agoritsas2018out,
  title={Out-of-equilibrium dynamical mean-field equations for the perceptron model},
  author={Agoritsas, Elisabeth and Biroli, Giulio and Urbani, Pierfrancesco and Zamponi, Francesco},
  journal={Journal of Physics A: Mathematical and Theoretical},
  volume={51},
  number={8},
  pages={085002},
  year={2018},
  publisher={IOP Publishing}
}

@article{biroli2005crash,
  title={A crash course on ageing},
  author={Biroli, Giulio},
  journal={Journal of Statistical Mechanics: Theory and Experiment},
  volume={2005},
  number={05},
  pages={P05014},
  year={2005},
  publisher={IOP Publishing}
}

@article{bouchaud1998out,
  title={Out of equilibrium dynamics in spin-glasses and other glassy systems},
  author={Bouchaud, Jean-Philippe and Cugliandolo, Leticia F and Kurchan, Jorge and Mezard, Marc},
  journal={Spin glasses and random fields},
  volume={12},
  pages={161},
  year={1998},
  publisher={World scientific Singapore}
}

@article{bray2002theory,
  title={Theory of phase-ordering kinetics},
  author={Bray, Alan J},
  journal={Advances in Physics},
  volume={51},
  number={2},
  pages={481--587},
  year={2002},
  publisher={Taylor \& Francis}
}

@article{arous2020algorithmic,
  title={Algorithmic thresholds for tensor PCA},
  author={Arous, Gerard Ben and Gheissari, Reza and Jagannath, Aukosh},
  journal={The Annals of Probability},
  volume={48},
  number={4},
  pages={2052--2087},
  year={2020},
  publisher={Institute of Mathematical Statistics}
}

@article{barbier2021finite,
  title={Finite size effects and loss of self-averageness in the relaxational dynamics of the spherical Sherrington-Kirkpatrick model},
  author={Barbier, Damien and Pimenta, Pedro H and Cugliandolo, Leticia F and Stariolo, Daniel A},
  journal={arXiv preprint arXiv:2103.12654},
  year={2021}
}

@article{cugliandolo1995full,
  title={Full dynamical solution for a spherical spin-glass model},
  author={Cugliandolo, Leticia F and Dean, David S},
  journal={Journal of Physics A: Mathematical and General},
  volume={28},
  number={15},
  pages={4213},
  year={1995},
  publisher={IOP Publishing}
}

@article{goldt2020dynamics,
  title={Dynamics of stochastic gradient descent for two-layer neural networks in the teacher--student setup},
  author={Goldt, Sebastian and Advani, Madhu S and Saxe, Andrew M and Krzakala, Florent and Zdeborov{\'a}, Lenka},
  journal={Journal of Statistical Mechanics: Theory and Experiment},
  volume={2020},
  number={12},
  pages={124010},
  year={2020},
  publisher={IOP Publishing}
}

@article{baik2005phase,
  title={Phase transition of the largest eigenvalue for nonnull complex sample covariance matrices},
  author={Baik, Jinho and Ben Arous, G{\'e}rard and P{\'e}ch{\'e}, Sandrine},
  journal={The Annals of Probability},
  volume={33},
  number={5},
  pages={1643--1697},
  year={2005},
  publisher={Institute of Mathematical Statistics}
}

@article{castellani2005spin,
  title={Spin-glass theory for pedestrians},
  author={Castellani, Tommaso and Cavagna, Andrea},
  journal={Journal of Statistical Mechanics: Theory and Experiment},
  volume={2005},
  number={05},
  pages={P05012},
  year={2005},
  publisher={IOP Publishing}
}

@article{thalmann2001geometrical,
  title={Geometrical approach for the mean-field dynamics of a particle in a short range correlated random potential},
  author={Thalmann, Fabrice},
  journal={The European Physical Journal B-Condensed Matter and Complex Systems},
  volume={19},
  number={1},
  pages={49--63},
  year={2001},
  publisher={Springer}
}

@article{mannelli2020marvels,
  title={Marvels and pitfalls of the langevin algorithm in noisy high-dimensional inference},
  author={Mannelli, Stefano Sarao and Biroli, Giulio and Cammarota, Chiara and Krzakala, Florent and Urbani, Pierfrancesco and Zdeborov{\'a}, Lenka},
  journal={Physical Review X},
  volume={10},
  number={1},
  pages={011057},
  year={2020},
  publisher={APS}
}

@article{mannelli2019afraid,
  title={Who is afraid of big bad minima? analysis of gradient-flow in spiked matrix-tensor models},
  author={Sarao Mannelli, Stefano and Biroli, Giulio and Cammarota, Chiara and Krzakala, Florent and Zdeborov{\'a}, Lenka},
  journal={Advances in Neural Information Processing Systems},
  volume={32},
  pages={8679--8689},
  year={2019}
}

@inproceedings{power2021grokking,
  title={Grokking: Generalization beyond overfitting on small algorithmic datasets},
  author={Power, Alethea and Burda, Yuri and Edwards, Harri and Babuschkin, Igor and Misra, Vedant},
  booktitle={ICLR MATH-AI Workshop},
  year={2021}
}

@article{crisanti1992spherical,
  title={The spherical p-spin interaction spin glass model: the statics},
  author={Crisanti, Andrea and Sommers, H-J},
  journal={Zeitschrift f{\"u}r Physik B Condensed Matter},
  volume={87},
  number={3},
  pages={341--354},
  year={1992},
  publisher={Springer}
}

@article{cugliandolo1993analytical,
  title={Analytical solution of the off-equilibrium dynamics of a long-range spin-glass model},
  author={Cugliandolo, Leticia F and Kurchan, Jorge},
  journal={Physical Review Letters},
  volume={71},
  number={1},
  pages={173},
  year={1993},
  publisher={APS}
}

@article{moulines2011non,
  title={Non-asymptotic analysis of stochastic approximation algorithms for machine learning},
  author={Moulines, Eric and Bach, Francis},
  journal={Advances in neural information processing systems},
  volume={24},
  pages={451--459},
  year={2011}
}

@article{xu2011towards,
  title={Towards optimal one pass large scale learning with averaged stochastic gradient descent},
  author={Xu, Wei},
  journal={arXiv preprint arXiv:1107.2490},
  year={2011}
}

@inproceedings{li2017stochastic,
  title={Stochastic modified equations and adaptive stochastic gradient algorithms},
  author={Li, Qianxiao and Tai, Cheng and Weinan, E},
  booktitle={International Conference on Machine Learning},
  pages={2101--2110},
  year={2017},
  organization={PMLR}
}

@article{liu2019deep,
  title={Deep learning theory review: An optimal control and dynamical systems perspective},
  author={Liu, Guan-Horng and Theodorou, Evangelos A},
  journal={arXiv preprint arXiv:1908.10920},
  year={2019}
}

@article{brea2019weight,
  title={Weight-space symmetry in deep networks gives rise to permutation saddles, connected by equal-loss valleys across the loss landscape},
  author={Brea, Johanni and Simsek, Berfin and Illing, Bernd and Gerstner, Wulfram},
  journal={arXiv preprint arXiv:1907.02911},
  year={2019}
}

@article{mannelli2020thresholds,
  title={Thresholds of descending algorithms in inference problems},
  author={Mannelli, Stefano Sarao and Zdeborov{\'a}, Lenka},
  journal={Journal of Statistical Mechanics: Theory and Experiment},
  volume={2020},
  number={3},
  pages={034004},
  year={2020},
  publisher={IOP Publishing}
}

@article{dembo2020dynamics,
  title={Dynamics for spherical spin glasses: disorder dependent initial conditions},
  author={Dembo, Amir and Subag, Eliran},
  journal={Journal of Statistical Physics},
  volume={181},
  number={2},
  pages={465--514},
  year={2020},
  publisher={Springer}
}

@inproceedings{cheng2020stochastic,
  title={Stochastic gradient and langevin processes},
  author={Cheng, Xiang and Yin, Dong and Bartlett, Peter and Jordan, Michael},
  booktitle={International Conference on Machine Learning},
  pages={1810--1819},
  year={2020},
  organization={PMLR}
}

@article{mingard2021sgd,
  title={Is SGD a Bayesian sampler? Well, almost},
  author={Mingard, Chris and Valle-P{\'e}rez, Guillermo and Skalse, Joar and Louis, Ard A},
  journal={Journal of Machine Learning Research},
  volume={22},
  number={79},
  pages={1--64},
  year={2021}
}

@article{mignacco2021effective,
  title={The effective noise of Stochastic Gradient Descent},
  author={Mignacco, Francesca and Urbani, Pierfrancesco},
  journal={arXiv preprint arXiv:2112.10852},
  year={2021}
}

@article{auffinger2013random,
  title={Random matrices and complexity of spin glasses},
  author={Auffinger, Antonio and Ben Arous, G{\'e}rard and {\v{C}}ern{\`y}, Ji{\v{r}}{\'\i}},
  journal={Communications on Pure and Applied Mathematics},
  volume={66},
  number={2},
  pages={165--201},
  year={2013},
  publisher={Wiley Online Library}
}

@article{hu2017diffusion,
  title={On the diffusion approximation of nonconvex stochastic gradient descent},
  author={Hu, Wenqing and Li, Chris Junchi and Li, Lei and Liu, Jian-Guo},
  journal={arXiv preprint arXiv:1705.07562},
  year={2017}
}

@article{zdeborova2016statistical,
  title={Statistical physics of inference: Thresholds and algorithms},
  author={Zdeborov{\'a}, Lenka and Krzakala, Florent},
  journal={Advances in Physics},
  volume={65},
  number={5},
  pages={453--552},
  year={2016},
  publisher={Taylor \& Francis}
}

@article{jastrzkebski2017three,
  title={Three factors influencing minima in sgd},
  author={Jastrzkbski, Stanislaw and Kenton, Zachary and Arpit, Devansh and Ballas, Nicolas and Fischer, Asja and Bengio, Yoshua and Storkey, Amos},
  journal={arXiv preprint arXiv:1711.04623},
  year={2017}
}

@inproceedings{park2019effect,
  title={The effect of network width on stochastic gradient descent and generalization: an empirical study},
  author={Park, Daniel and Sohl-Dickstein, Jascha and Le, Quoc and Smith, Samuel},
  booktitle={International Conference on Machine Learning},
  pages={5042--5051},
  year={2019},
  organization={PMLR}
}

@article{smith2017don,
  title={Don't decay the learning rate, increase the batch size},
  author={Smith, Samuel L and Kindermans, Pieter-Jan and Ying, Chris and Le, Quoc V},
  journal={arXiv preprint arXiv:1711.00489},
  year={2017}
}

@article{ros2019complex,
  title={Complex energy landscapes in spiked-tensor and simple glassy models: Ruggedness, arrangements of local minima, and phase transitions},
  author={Ros, Valentina and Ben Arous, Gerard and Biroli, Giulio and Cammarota, Chiara},
  journal={Physical Review X},
  volume={9},
  number={1},
  pages={011003},
  year={2019},
  publisher={APS}
}

@article{arous2019landscape,
  title={The landscape of the spiked tensor model},
  author={Ben Arous, Gerard and Mei, Song and Montanari, Andrea and Nica, Mihai},
  journal={Communications on Pure and Applied Mathematics},
  volume={72},
  number={11},
  pages={2282--2330},
  year={2019},
  publisher={Wiley Online Library}
}

@article{sherrington1975solvable,
  title={Solvable model of a spin-glass},
  author={Sherrington, David and Kirkpatrick, Scott},
  journal={Physical review letters},
  volume={35},
  number={26},
  pages={1792},
  year={1975},
  publisher={APS}
}

@article{wigner1958distribution,
  title={On the distribution of the roots of certain symmetric matrices},
  author={Wigner, Eugene P},
  journal={Annals of Mathematics},
  pages={325--327},
  year={1958},
  publisher={JSTOR}
}

@article{gilmer2021loss,
  title={A Loss Curvature Perspective on Training Instability in Deep Learning},
  author={Gilmer, Justin and Ghorbani, Behrooz and Garg, Ankush and Kudugunta, Sneha and Neyshabur, Behnam and Cardoze, David and Dahl, George and Nado, Zachary and Firat, Orhan},
  journal={arXiv preprint arXiv:2110.04369},
  year={2021}
}

@article{gotmare2018closer,
  title={A closer look at deep learning heuristics: Learning rate restarts, warmup and distillation},
  author={Gotmare, Akhilesh and Keskar, Nitish Shirish and Xiong, Caiming and Socher, Richard},
  journal={arXiv preprint arXiv:1810.13243},
  year={2018}
}

@article{goyal2017accurate,
  title={Accurate, large minibatch sgd: Training imagenet in 1 hour},
  author={Goyal, Priya and Doll{\'a}r, Piotr and Girshick, Ross and Noordhuis, Pieter and Wesolowski, Lukasz and Kyrola, Aapo and Tulloch, Andrew and Jia, Yangqing and He, Kaiming},
  journal={arXiv preprint arXiv:1706.02677},
  year={2017}
}

@article{you2019does,
  title={How does learning rate decay help modern neural networks?},
  author={You, Kaichao and Long, Mingsheng and Wang, Jianmin and Jordan, Michael I},
  journal={arXiv preprint arXiv:1908.01878},
  year={2019}
}

@article{li2019towards,
  title={Towards explaining the regularization effect of initial large learning rate in training neural networks},
  author={Li, Yuanzhi and Wei, Colin and Ma, Tengyu},
  journal={arXiv preprint arXiv:1907.04595},
  year={2019}
}

@inproceedings{smith2017cyclical,
  title={Cyclical learning rates for training neural networks},
  author={Smith, Leslie N},
  booktitle={2017 IEEE winter conference on applications of computer vision (WACV)},
  pages={464--472},
  year={2017},
  organization={IEEE}
}

@article{loshchilov2016sgdr,
  title={Sgdr: Stochastic gradient descent with warm restarts},
  author={Loshchilov, Ilya and Hutter, Frank},
  journal={arXiv preprint arXiv:1608.03983},
  year={2016}
}

@article{lewkowycz2021decay,
  title={How to decay your learning rate},
  author={Lewkowycz, Aitor},
  journal={arXiv preprint arXiv:2103.12682},
  year={2021}
}

@article{duchi2011adaptive,
  title={Adaptive subgradient methods for online learning and stochastic optimization.},
  author={Duchi, John and Hazan, Elad and Singer, Yoram},
  journal={Journal of machine learning research},
  volume={12},
  number={7},
  year={2011}
}

@article{zeiler2012adadelta,
  title={Adadelta: an adaptive learning rate method},
  author={Zeiler, Matthew D},
  journal={arXiv preprint arXiv:1212.5701},
  year={2012}
}

@article{kingma2014adam,
  title={Adam: A method for stochastic optimization},
  author={Kingma, Diederik P and Ba, Jimmy},
  journal={arXiv preprint arXiv:1412.6980},
  year={2014}
}

@article{keskar2017improving,
  title={Improving generalization performance by switching from adam to sgd},
  author={Keskar, Nitish Shirish and Socher, Richard},
  journal={arXiv preprint arXiv:1712.07628},
  year={2017}
}

@article{chen2018closing,
  title={Closing the generalization gap of adaptive gradient methods in training deep neural networks},
  author={Chen, Jinghui and Zhou, Dongruo and Tang, Yiqi and Yang, Ziyan and Cao, Yuan and Gu, Quanquan},
  journal={arXiv preprint arXiv:1806.06763},
  year={2018}
}

@article{wilson2017marginal,
  title={The marginal value of adaptive gradient methods in machine learning},
  author={Wilson, Ashia C and Roelofs, Rebecca and Stern, Mitchell and Srebro, Nathan and Recht, Benjamin},
  journal={arXiv preprint arXiv:1705.08292},
  year={2017}
}

@article{ge2019step,
  title={The step decay schedule: A near optimal, geometrically decaying learning rate procedure for least squares},
  author={Ge, Rong and Kakade, Sham M and Kidambi, Rahul and Netrapalli, Praneeth},
  journal={arXiv preprint arXiv:1904.12838},
  year={2019}
}

@article{arous2020classification,
  title={A classification for the performance of online SGD for high-dimensional inference},
  author={Ben Arous, Gerard and Gheissari, Reza and Jagannath, Aukosh},
  journal={arXiv:2003.10409},
  year={2020}
}

@inproceedings{bottou2003stochastic,
  title={Stochastic learning},
  author={Bottou, L{\'e}on},
  booktitle={Summer School on Machine Learning},
  pages={146--168},
  year={2003},
  organization={Springer}
}

@inproceedings{refinetti2021align,
  title={Align, then memorise: the dynamics of learning with feedback alignment},
  author={Refinetti, Maria and d’Ascoli, St{\'e}phane and Ohana, Ruben and Goldt, Sebastian},
  booktitle={International Conference on Machine Learning},
  pages={8925--8935},
  year={2021},
  organization={PMLR}
}

@inproceedings{choromanska2015loss,
  title={The loss surfaces of multilayer networks},
  author={Choromanska, Anna and Henaff, Mikael and Mathieu, Michael and Ben Arous, G{\'e}rard and LeCun, Yann},
  booktitle={Artificial intelligence and statistics},
  pages={192--204},
  year={2015},
  organization={PMLR}
}

\newpage
\appendix
\onecolumn
\numberwithin{equation}{section}

\section{Dynamics of the convex model}
\label{app:convex}
Here we give additional details and steps in the computations on the convex model of Sec.~\ref{sec:convex}.
The loss function is given by $\loss(\spin) = \frac{\kappa}{2} \spin^2$. Integrating the Langevin equation (Eq.~\ref{eq:langevin}) from $t_0$ to $t$ for $\spin$ yields:
\begin{align}
    \spin(t) = \underbrace{\spin(t_0) e^{-\kappa \int_{t_{0}}^t \dd \tau \lr(\tau) }}_{\bar \spin(t)} + \underbrace{\int_{t_0}^t \dd t' e^{-\kappa \int_{t_0}^{t'} \dd t\tau \lr(\tau) }\lr(t') \noise(t')}_{\delta \spin(t)}.
\end{align}
In order to obtain a typical realisation of the loss which does not depend on the optimisation noise $\noise$, we take the expectation over $\xi$.
This gives for the loss $\loss$:
\begin{align}
   \langle \loss(t) \rangle &= \frac{\kappa}{2}\left(\langle \bar \spin(t)^2 \rangle + \langle \delta \spin(t)^2 \rangle + 2 \underbrace{\langle \bar \spin(t) \delta \spin(t) \rangle}_{0} \right) \\ 
    &= \frac{\kappa}{2}\left(\underbrace{\spin(t_0)^2 e^{-2\kappa \int_{t_0}^t \dd \tau \lr(\tau)}}_{\bar \loss(t)} + \underbrace{2T \int_{t_0}^t \dd t' \lr(t')^2 e^{-2\kappa\int_{t'}^{t} \dd \tau \lr(\tau)}}_{\delta \loss(t)} \right)
\end{align}
The first term is an optimisation term while the second is the contribution of the noise inherent to the optimisation algorithm. Thus, to converge to the solution as quickly as possible, one has to find the trade-off between decreasing the impact of the noise term while not slowing down optimisation excessively. The ideal schedule is determined by requiring these two effects are comparable. Defining $\lr(t)=\lr_0/t$, we obtain 
\begin{align}
    \bar \loss(t) &\propto e^{-2 \lr_0 \kappa \log(t)} \propto t^{- 2\lr_0\kappa} \\
    \delta \loss(t) & \int_{t_0}^t \dd t' \frac{1}{t'^2} \left(\frac{t'}{t}\right)^{2\lr_0\kappa}  \propto 1 / t.
\end{align}
If $\eta_0>\nicefrac{1}{2\kappa}$, the loss is dominated by the noise term $\delta\mathcal L$ and decays as $1/t$. If $\eta_0<\nicefrac{1}{2\kappa}$, the loss is dominated by the optimization term $\delta\mathcal L$ and decays as $t^{-2\eta_0\kappa}$.

\section{Dynamics of the Sherrington-Kirkpatrick model}
\label{app:SK}

In this section, we provide derivations for the results obtained in the SK model.

\subsection{Unplanted model}

The loss function is given by:
\begin{align}
    \begin{split}
        \loss(\spin) = -\frac{1}{\sqrt{N}} \sum_{i<j}^N J_{i j} \spin_{i} \spin_{j}. 
    \end{split}
\end{align}

\paragraph{Solving the dynamics}

Following~\citet{cugliandolo1995full}, we express the spin configurations in the eigenbasis of $\coupling$ and define $\spin_\mu = \nicefrac{\spin\cdot \coupling_\mu}{\sqrt N}$ as the projection of $\spin$ onto the eigenvector $\coupling_\mu$. $x_\mu$ evolves as:
\begin{align}
    \frac{\partial \spin_{\mu}(t)}{\partial t}&= \lr(t) \left[ (\mu-\spherical(t)) \spin_{\mu}(t)+\noise_{\mu}(t)\right].
\end{align}
Integrating this equation yields again two terms, one related to the optimisation and the second related to the noise:
\begin{align}
    \spin_{\mu}(t)&= \spin_{\mu}\left(0\right) e^{-\int_{0}^{t} d \tau \lr(\tau)(\mu-\spherical(\tau))}\\\notag
    &+\int_{0}^{t} d t^{\prime \prime} e^{-\int_{t^{\prime \prime}}^{t} d \tau^{\prime}\lr(\tau^\prime) (\spherical(\tau^\prime)-\mu)}\lr(t^{\prime\prime})\noise_{\mu}\left(t^{\prime \prime}\right).
\end{align}

In the $t\to\infty$ limit, a non-exploding $\bar \spin_\mu$ requires $\mu\!-\!\spherical(t)$ to be negative for all $\mu$ in the support of $\rho$, implying $\spherical(t)<2$.
We must also impose $\spherical(t)\to_{t\to\infty} 2$, otherwise $\spin_\mu(t)\to 0 \ \forall \mu$, in contradiction with the spherical constraint. 
To comply with these two requirements we define $\spherical(t) = 2-f(t)$, with $f(t)\to_{t\to\infty}0$.

In the constant learning rate setup $\lr(t)=1$ we know from ~\cite{cugliandolo1995full} that $f(t) = 3/(4t)$. With $\lr(t)=\lr_0/t^\lrexponent$, a natural ansatz is $f(t) = c/t^{1-\lrexponent}$. To determine $c$, we impose the spherical constraint:
\begin{align*}
    1 &= \langle \int \frac{\dd \mu}{N} \rho (\mu) \spin_\mu(t)^2 \rangle\\
    &= t^{2c\lr_0} \int_{-2}^{2} \dd \mu \sqrt{4-\mu^2} e^{2\lr_0 (\mu-2) t^{1-\lrexponent}} \\
    &= t^{2c\lr_0 - 3(1-\lrexponent)/2} \int_0^\infty \dd \epsilon \sqrt{2\epsilon} e^{-2\lr_0\epsilon}\propto t^{2c\lr_0 - 3(1-\lrexponent)/2}\notag
\Rightarrow c =\frac{3(1-\lrexponent)}{4\lr_0}.
\end{align*}

For $\lrexponent=1$, we instead use the ansatz $\spherical(t) = 2-c/\log(t)$:
\begin{align*}
    1 &= \int_{-2}^{2} \dd \mu \sqrt{4-\mu^2} e^{2\lr_0 (\mu-2) \log t} e^{2c\lr_0\log\log t} \\
    &= (\log t)^{2c\lr_0 - 3/2} \int_0^\infty \dd \epsilon \sqrt{2\epsilon} e^{-2\lr_0\epsilon}\propto  (\log t)^{2 c\lr_0 -3/2}
\Rightarrow c =\frac{3}{4\lr_0}.
\end{align*}

Hence, the scaled loss $\ell = \nicefrac{\loss}{N}$ converges to the ground state (global minimum) $\ell_{GS}\!~=~\!-1$ as a sum of power-laws:
\begin{align}
    \ell(t) - \ell_{GS} &= \frac{\lr_0 T}{2t^\lrexponent} + \begin{cases}
    \frac{3(1-\lrexponent)}{8 \lr_0 t^{1-\lrexponent}}, \quad \beta<1\\
    \frac{3}{8 \lr_0 \log t}, \quad \beta=1
    \end{cases}.
\end{align}

\paragraph{Dependency on the spectrum of $\coupling$}
One may naturally ask whether our conclusions are affected by changing the spectrum of the coupling matrix $\coupling$. Notice that the key to solving the self-consistent equation is the behavior of the spectrum near its right edge. For the semi-circle law considered here, the right edge of the spectrum behaves as a square root. This law applies to a rather wide range of random matrix ensembles. Besides, many other common spectral densities, such as the Marcenko-Pastur law, also exhibit a same square root behavior on their right edge. Hence we expect our results to hold for a wide range of random matrix ensembles.

\subsection{Planted model}

The loss function is given by:
\begin{align}
    \begin{split}
        \loss(\spin) &= - \frac{N}{2} m^2-\frac{\Delta}{\sqrt{N}} \sum_{i<j}^N \coupling_{i j} \spin_{i} \spin_{j}.
    \end{split}
\end{align}

\paragraph{Solving the dynamics}

Again we choose $\lr(t)=\lr_0/t^\lrexponent$ and consider the high signal-to-noise setting, $\Delta<\frac{1}{2}$. Writing $\spherical(t) = 1 - f(t)$, we obtain:
\begin{align}
    1 = &\langle \int\frac{\dd \mu}{N} \rho (\mu) \spin_\mu(t)^2 \rangle \\
    = &\frac{N-1}{N}\int_{-2}^{2} \dd \mu \rho_{sc}(\mu) e^{2\int_{t_0}^t \dd \tau \lr(\tau) (\mu-1) } e^{2\int_{t_0}^t \dd \tau \lr(\tau) f(t)} + \frac{1}{N} e^{2\int_{t_0}^t \dd \tau \lr(\tau) f(t)}\\
    = & e^{2\int_{t_0}^t\dd \tau \lr(\tau) f(\tau)} \left( \underbrace{e^{-2\lr(t)(1-2\Delta) }\int_{-2}^{2} \dd\mu \rho_{sc}(\mu/\Delta) e^{2\lr(t)(\mu-2\Delta) t}}_{A(t)} + \frac{1}{N}\right)
    \label{eq:competition}
\end{align}

The expression above involves two terms. The first is of order one but decays exponentially over time; using results above, we obtain that
\begin{align}
    A(t)\sim
    t^{-3(1-\lrexponent)/2}e^{-2\lr_0\kappa t^{1-\lrexponent}}.
\end{align}
Hence, there is a crossover time at which the first term becomes smaller than the second term, given by:
\begin{align}
    A(t)\sim1 \Rightarrow \tjump = 
    \left( \frac{\log N}{2\lr_0 \kappa }\right)^{\frac{1}{1-\lrexponent}}
\end{align}

Before $\tjump$, the signal is not detected and we have as before $\spherical(t) = 2\Delta-c/ t^{1-\lrexponent}$. 

After $\tjump$, we have $A(t)\ll 1/N$. Multiply Eq.~\ref{eq:competition} by $N$ and taking the log, we obtain:
\begin{align}
    \log N &= 2 \int_{t_0}^t\dd \tau f(\tau) t^{-\lrexponent} + \log\left( 1 + N A(t)\right)\\
    &\sim 2 \int_{t_0}^t\dd \tau f(\tau) t^{-\lrexponent} + N A(t)
\end{align}

Taking the derivative with respect to $t$, we find the following asymptotics for late times:
\begin{align}
    f(t) \sim \frac{-NA'(t)t^\lrexponent}{2} \sim t^{-5(1-\lrexponent)/2} e^{-2\lr_0\kappa t^{1-\lrexponent}}
\end{align}

Hence,
\begin{align}
    \ell(t)-\ell_{GS} = \frac{\lr_0 T}{2t^{\lrexponent}} + \frac{1}{2} f(t)
\end{align}

with $\ell_{GS}=1$. As previously, it is straightfoward to extend this to the setup $\beta=1$, for which we obtain $f(t)\sim t^{-2\eta_0\kappa}$.

\paragraph{Curvature analysis}

\begin{figure}
    \centering
    \includegraphics[width=.5\columnwidth]{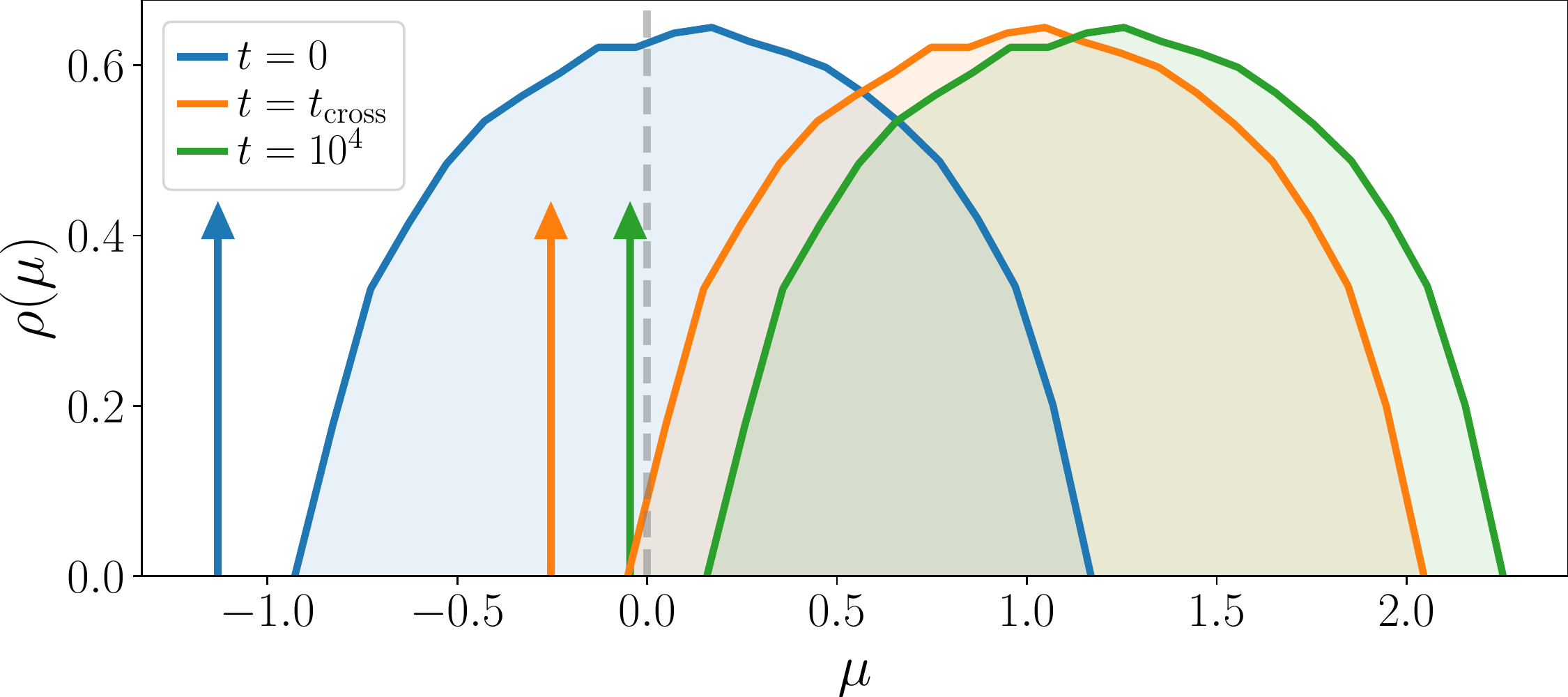}
    \caption{\textbf{The landscape becomes convex at the crossover time.} \emph{Parameters: $N=3000$, $\lr_0= 0.1$}.}
    \label{fig:hessian-planted}
\end{figure}

As before, the spectrum of interest to study the landscape is that of $H$ shifted to the right by the spherical constraint $\spherical(t)$, depicted in Fig.~\ref{fig:hessian-planted}. 
The crossover time $\tjump$ corresponds to the time at which the left edge of the semi-circle reaches 0. Thanks to the presence of the signal, the dynamics do not stop at this point; they continue until the eigenvalue corresponding to the signal reaches zero (which as achieved at $t\to\infty$). After the $\tjump$, the landscape becomes locally convex: the only negative eigenvalue is in the direction of the signal. Due to the spherical constraint, the effective Hessian (of dimension $N-1$) does not feel this negative eigenvalue when $\spin$ is close to $\signal$.

\section{Dynamics of the p-spin model}
\label{app:pspin}

\subsection{Rescaling the temperature}
\label{app:general_scaling}
Introducing a learning rate schedule is equivalnt to changing the "clock" directly in the Langevin (Eq.~\ref{eq:langevin}) as $\dd\tilde{t}=\lr(t)\dd t$. Then, for $\beta<1$ we have:
\begin{align}
\begin{split}
    \delta(\tilde{t} ) \dd \tilde{t}  =\delta(t) \dd t 
    \Rightarrow \delta(t) & =
    \delta(\tilde{t}) \left(\frac{\lr_0}{1-\beta}\right)^{\frac{\beta}{1-\beta}}\tilde{t}^{\frac{-\beta}{1-\beta}}
\end{split}
\end{align}

The Langevin equation becomes:
\begin{align}
    \begin{split}
        \frac{\dd \spin_i(t)}{\dd t} = -\left(\frac{\partial \loss(\spin, \signal)}{\partial \spin_i} + \noise_i(t) + \spherical(t)\spin_i(t) \right),
        \qquad \langle\noise(t)\noise(t^\prime)\rangle = 2 T\tilde{\lr}_0\tilde{t}^{\nicefrac{-\lrexponent}{1-\lrexponent}}  \delta(t-t'),
    \end{split}
    \label{eq:langevin2}
\end{align}
where we defined $\tilde{\lr}_0 = \left(\frac{\lr_0}{1-\beta}\right)^{\frac{\beta}{1-\beta}}$.
This equation reveals that the process optimised with a varying learning rate is equivalent to a process at an effective temperature:
\begin{equation}
    \tilde{T} = \tilde{\lr}_0\tilde{t}^{\frac{-\lrexponent}{1-\lrexponent}} T
\end{equation}
This equation corresponds to the physical protocol in which the temperature $\tilde{T}$ is annealed as a power law.
As we show for the $p$-spin model in the next section, the solution is governed by a speed-noise trade-off. 

\subsection{Application to the p-spin model}
For the $p$-spin model the loss can therefore be written:
\begin{equation}
\loss(t; \tilde{T}) =\frac{-N}{p}\left(\tilde{z}(t) - \tilde{T}\right) = 
    \loss(t; T=0) + \loss_\th(\tilde{T})
\end{equation}
where we assumed that, at all times, the temperature dependent contribution to the loss has time to equilibrate in the threshold states.

To find $\loss_\th(\tilde{T})$, we assume that, since we are looking at long times, we have $\tilde{T}\ll 1$. We can consider the loss by performing an expansion around the $\tilde T=0$ minimum i.e. considering that the motion is oscillatory around the minimum. At $T=0$, the threshold overlap is given by $q_{\th}=1$. At $T\ll 1$, we thus write $q = 1- \chi T$. Performing a similar matching argument as the one of \cite{mannelli2020marvels}, described in more details in Sec.~\ref{sec:matching}, we find that the close to the threshold, the loss is given by Eq.~\ref{eq:eth_SMTM}:
\begin{align}
\begin{split}
\label{eq:Lossth1}
    \loss_\th(T) &= -\frac{1}{p}\left[\sqrt{(p-1)q^{p-2}} + \frac{1}{ T}\sqrt{\frac{2}{p}}\left(1 - q^{p-1}\right)\right]
    \end{split}
\end{align}
In addition, we can expand the threshold overlap solution around $T=0$ \cite{mannelli2019afraid}:
\begin{align}
    \begin{split}
        q_\th^{p-2} (1-q_\th)^2 &= \frac{T^2}{p-1}\\
        \Rightarrow \chi = \sqrt{\frac{1}{(p-1)}}
    \end{split}
\end{align}
Replacing this solution in Eq.~\ref{eq:Lossth1}, we find:

\begin{align}
    \begin{split}
        \loss_\th(T) &= \underbrace{ - \frac{\sqrt{4(p-1)}}{p}}_{\loss_\th(T=0)} + \underbrace{\frac{p-2}{p} T}_{\propto T}
    \end{split}
\end{align}

We see that $\loss_\th(T)$ is composed of a constant term, which is the same as the threshold loss defined in Eq.~\ref{eq:loss_threshold} and a term scaling linearly with $T$. Thus:
\begin{align}
    \begin{split}
        \loss_\th(\tilde T) - \loss_\th \propto \tilde T \propto t^{- \nicefrac{\lrexponent}{1-\lrexponent}}.
    \end{split}
\end{align}
We find again the two competing term in the speed of optimisation. 
On the one hand, the noiseless term, which is the same as the zero temperature loss, decays as $\loss(t; T=0)\propto t^{-\gamma}$. On the other hand, the temperature dependent term, which blocks the dynamics at loss $\loss_\th(\tilde T)$, which decays as $t^{-\nicefrac{\lrexponent}{1-\lrexponent}}$. The loss decays as $t^{-\mathrm{min}\left(\gamma, \frac{\lrexponent}{1-\lrexponent}\right)}$. Equaling the two exponents gives the optimal value of $\lrexponent_\opt=\frac{2}{5}$.


\section{Dynamics of the Spiked Matrix-Tensor model}
\label{app:SMTM}
\subsection{Derivation of the PDE equations}
\label{app:derivationPDE}
For simplicity, we detail the derivation of the \emph{p-spin model} without the spike as studied in Sec.~\ref{sec:pspinmodel} in the case $p=3$. The derivation for the full spiked tensor model is similar and can be found in \cite{mannelli2019afraid}.
The Langevin equation for each spin $\spin$ is given by: 
\begin{equation}
    \dot{\spin}_i(t)=-\spherical(t)\lr(t) - \lr(t)\partial_{\spin_i}\loss + \lr(t)\noise_i(t) 
\end{equation}
where $\bnoise\in\R^N$ is the Langevin noise with distribution $\langle \noise_i(t)\rangle=0$ and $\langle \noise_i(t)\noise_j(t')\rangle=2T\delta_{ij}\delta(t-t')$. The solution $\spin$ to the Langevin equation depends on the realisation of the noise. We can obtain a probability distribution over $\spin$ given the distribution of $\noise$ by considering the expectation of an observable $A(t)$:
\begin{align}
    \langle A(x)\rangle=\int D \noise P(\noise) A\left(\spin_{\noise}\right) =\int d x\left[\int D \noise P(\noise) \delta\left(\dot x+\lr(t)\partial_{x} \loss-\lr(t)\noise\right)\right] A(x) =\int d x P(x) A(x)
\end{align}
We are now interested in considering $P(x)$ when averaged over the quenched disorder $J$. We therefore resort to:
\begin{align}
\begin{split}
    1 \equiv Z 
    &=\int D x P(x)\\
&=\int D x D \hat{\spin} D \noise \exp \left[-\frac{1}{2} \int d t d t^{\prime} \noise(t) D_0^{-1}\left(t, t^{\prime}\right) \noise\left(t^{\prime}\right)+i \int d t \hat{\spin}(t)\left(\partial_{t} x+\lr(t)\partial_{x} \loss\right)-i \int d t \lr(t)\hat{\spin}(t) \noise(t)\right]\\
&= \int D x D \hat{\spin} \exp \left[
-\frac{1}{2} \int d t d t^{\prime} \hat x(t)\lr(t) D_0\left(t, t^{\prime}\right) \hat x\left(t^{\prime}\right)\lr(t')+i \int d t \hat{\spin}(t)\left(\partial_{t} x+\lr(t)\partial_{x} \loss\right)\right]\\
&= \int D x D \hat{\spin} \exp \left[S(x, \hat x)\right]
\end{split}
\end{align}
where we defined $D_0(t, t') =2T\delta(t-t')$. Crucially, $S(x,\hat x)$ acts as a generating functional and allows to obtain correlation functions by a term $\int d t \hat x(t) h(t) + x(t) \hat h(t)$. We can thus define
\begin{align}
    \langle x(t)\hat x(t') \rangle = \partial_{h(t')} \langle x(t)\rangle \equiv R(t, t') \qquad
    \langle x(t) x(t') \rangle = \partial_{\hat h(t')}\langle x(t)\rangle\equiv C(t, t')
\end{align}
We now want to average the partition function over the quenched disorder $Z$. We note that the only time depend term in the exponent is $i \hat x(t) \lr(t) \partial_x \loss$. We thus have to compute:
\begin{align}
    \begin{split}
      \overline{  e^{i \hat x(t) \lr(t) \partial_x \loss  }} \equiv e^{\Delta(x, \hat x)} 
    \end{split}
\end{align}
The average over the disorder will induce corrections both to the propagator $D_0$ and to the interaction term $\hat x(t) x(t')$ i.e. $\Delta(x, \hat{\spin})=-\frac{1}{2} \hat{\spin} D_{1}(x, \hat{\spin}) \hat{\spin}+i \hat{\spin} \mathcal{L}_{1}(x, \hat{\spin})$. 
By performing the average we obtain: 
\begin{align}
    \begin{split}
\overline{i \hat \spin_i(t) \lr \partial_{\spin_i}  \loss  }&=\int \prod_{i>k>l} d \coupling_{i k l} \exp \left\{-\frac{1}{2} \coupling_{i k l}^{2}-\sqrt{\frac{(p-1)!}{N^{p-1}}} \coupling_{i k l} \int d t\lr(t)\left[i \hat{\spin}_{i} \spin_{k} \spin_{l}+\spin_{i} i \hat{\spin}_{k} \spin_{l}+\spin_{i} \spin_{k} i \hat{\spin}_{l}\right]\right\}\\
&=\exp \left\{\int \frac{d t d t^{\prime}}{2 N^{p-1}}\lr(t)\lr(t')\left[(i \hat{\spin} \cdot i \hat{\spin})(x \cdot x)^{p-1}+(p-1)(i \hat{\spin} \cdot x)(x \cdot i \hat{\spin})(x \cdot x)^{p-2}\right]\right\}
    \end{split}
\end{align}
where we introduced the notation $x \cdot x \equiv \sum_{i=1}^{N} \spin_{i}(t) \spin_{i}\left(t^{\prime}\right)$.
We now introduce dynamical overlaps $Q_1$, $Q_2$, $Q_3$ and $Q_4$ as:
\begin{align}
    \begin{split}
 \overline{e^{i \hat \spin_i(t) \lr \partial_{\spin_i}  \loss } }
=& \int D Q \delta\left(N Q_{1}-\sum_{k} i \hat{\spin}_{k}(t) i \hat{\spin}_{k}\left(t^{\prime}\right)\right) \delta\left(N Q_{2}-\sum_{k} \spin_{k}(t) \spin_{k}\left(t^{\prime}\right)\right) \\
\cdot & \delta\left(N Q_{3}-\sum_{k} i \hat{\spin}_{k}(t) \spin_{k}\left(t^{\prime}\right)\right) \delta\left(N Q_{4}-\sum_{k} \spin_{k}(t) i \hat{\spin}_{k}\left(t^{\prime}\right)\right) \\
\cdot & \exp \left\{\frac{N}{2} \int d t d t^{\prime}\lr(t)\lr(t')\left[Q_{1}\left(t, t^{\prime}\right) Q_{2}\left(t, t^{\prime}\right)^{p-1}+(p-1) Q_{3}\left(t, t^{\prime}\right) Q_{4}\left(t, t^{\prime}\right) Q_{2}\left(t, t^{\prime}\right)^{p-2}\right]\right\}
    \end{split}
\end{align}
We can easily see that we have the correspondence $Q_1(t, t')=0$, $Q_2(t, t')=C(t, t')$, $Q_3(t, t')=R(t', t)$ and $Q_4(t, t')=R(t, t')$.
By using the exponential form of the delta function and solving the fix point equations for the conjugate fields $\hat Q_1$, $\hat Q_2$, $\hat Q_3$ and $\hat Q_4$ we find:
\begin{align}
    \begin{cases}
i\hat Q_{1}=\frac{1}{2}\lr(t)\lr(t') Q_{2}^{p-1} \\
i\hat Q_{2}=\frac{p-1}{2}\lr(t)\lr(t') Q_{1} Q_{2}^{p-2}+\frac{(p-1)(p-2)}{2}\lr(t)\lr(t') Q_{3} Q_{4} Q_{2}^{p-3} \equiv 0 \\
i\hat Q_{3}=\frac{p-1}{2}\lr(t)\lr(t') Q_{4} Q_{2}^{p-2} \\
i\hat Q_{4}=\frac{p-1}{2}\lr(t)\lr(t') Q_{3} Q_{2}^{p-2}
\end{cases}
\end{align}
From the definition of the $\hat Q$'s we find the new term in the generating functional as:
\begin{equation}
    \Delta=\sum_{k} \int d t d t^{\prime}\lr(t)\lr(t')\left\{-\frac{1}{2} C\left(t, t^{\prime}\right)^{p-1} \hat{\spin}_{k}(t) \hat{\spin}_{k}\left(t^{\prime}\right)- (p-1) R\left(t, t^{\prime}\right) C\left(t, t^{\prime}\right)^{p-2} i \hat{\spin}_{k}(t) \spin_{k}\left(t^{\prime}\right)\right\}
\end{equation}
This allows us to write an effective Langevin equation for a scalar degree of freedom $\spin$:
\begin{equation}
    \dot x(t) = -\spherical(t)\lr(t)x(t) +  \lr(t)(p-1)\int d t^{\prime \prime}\lr(t') R\left(t, t^{\prime \prime}\right) C\left(t, t^{\prime \prime}\right)^{p-2} \sigma\left(t^{\prime \prime}\right)+\lr(t)\tilde \noise(t),
\end{equation}
with:
\begin{equation}
    \langle \tilde \noise(t)\tilde \noise(t') \rangle = 2 T \delta(t-t') + C^{p-1}(t, t').
\end{equation}
In order to write down a set of PDE's for $R$ and $C$, note the useful relations:
\begin{align}
\begin{split}
    \langle\frac{\partial x(t)}{\partial\noise(t')}\rangle &= -i  \langle x(t) \hat x(t')\rangle\\
    \langle x(t)\noise(t')\rangle &=  2T\lr(t') R(t, t^{'}) \\
     \langle \tilde \noise(t_1)x(t_2)\rangle &= 2T\lr(t_1) R(t_1, t_2) +  \int dt^{\prime\prime} \lr(t^{\prime\prime}) R(t^{\prime\prime}, t_2) C^{p-1}(t^{\prime\prime}, t_1)
\end{split}    
\end{align}
We therefore find:
\begin{tcolorbox}[ams align]
\label{eq:dynamical_equations_no_signal_R}
    \begin{split}
\frac{\partial R\left(t_{1}, t_{2}\right)}{\partial t_{1}}=&\left\langle\frac{\delta \dot{x}\left(t_{1}\right)}{\delta \tilde \noise\left(t_{2}\right)}\right\rangle\\
=&-z\left(t_{1}\right)\lr(t_1) R\left(t_{1}, t_{2}\right)+ \lr(t_1)\delta\left(t_{1}, t_{2}\right)\\
& + (p-1)  \lr(t_1)\int_{t_{2}}^{t_{1}} d t^{\prime \prime} \lr(t^{\prime\prime}) R\left(t_{1}, t^{\prime \prime}\right) C^{p-2}\left(t_{1}, t^{\prime \prime}\right) R\left(t^{\prime \prime}, t_{2}\right)
\end{split}
\end{tcolorbox}

\begin{tcolorbox}[ams align]
\label{eq:dynamical_equations_no_signal_C}
\begin{split}
\frac{\partial C\left(t_{1}, t_{2}\right)}{\partial t_{1}}=&\left\langle\dot{x}\left(t_{1}\right) x\left(t_{2}\right)\right\rangle\\
=& -\lr(t_1)z\left(t_{1}\right) C\left(t_{1}, t_{2}\right)+ 2T\lr(t_1)^2 R(t_1, t_2)\\
&+ (p-1) \lr(t_1) \int_{-\infty}^{t_{1}} d t^{\prime \prime} \lr(t^{\prime\prime})R\left(t_{1}, t^{\prime \prime}\right) C^{p-2}\left(t_{1}, t^{\prime \prime}\right) C\left(t^{\prime \prime}, t_{2}\right) \\
&+ \lr(t_1) \int dt^{\prime\prime} \lr(t^{\prime\prime}) R(t^{\prime\prime}, t_2) C^{p-1}(t^{\prime\prime}, t_1)\\
\end{split}
\end{tcolorbox}

The equation for $\spherical(t)$ is given by differentiation $C(1,1)=1$, i.e. $\left[\partial_{t} C\left(t, t^{\prime}\right)+\partial_{t^{\prime}} C\left(t, t^{\prime}\right)\right]_{t, t^{\prime}=s}=0$:
\begin{tcolorbox}[ams align]
\label{eq:dynamical_equations_no_signal_spherical}
\begin{split}
&\spherical(t_1) = T\lr(t_1) + p\int dt_2 \lr(t_2) R(t_2, t_1) C^{p-1}(t_2, t_1).
\end{split}
\end{tcolorbox}
The loss at all times is found by using the Ito identity: 
\begin{equation}
    \frac{1}{N} \frac{d}{d t} \sum_{i} \spin_{i}^{2}(t)=\frac{2}{N} \sum_{i} \spin_{i}(t) \dot{x}_{i}(t)+2
\end{equation}
which yields:
\begin{equation}
    \loss(t) = \frac{N}{p}\left(  T \lr(t) - \spherical(t)\right)
\end{equation}

\paragraph{Spiked matrix-tensor model}
The derivation of the PDEs describing the dynamics of $C$, $R$ and $z$ in the spiked matrix-tensor model are similar as the ones for the $p$-spin. In addition, one also needs to keep track of the evolution of the overlap of the estimate with the signal i.e. the magnetisation $m=\nicefrac{\spin\cdot\signal}{N}$. Using the same method as before we find:

\begin{tcolorbox}[ams align]
    \begin{split}
\frac{\partial}{\partial t} C\left(t, t^{\prime}\right)=&-\spherical(t)\lr(t) C\left(t, t^{\prime}\right)+\lr(t)Q^{\prime}(m(t)) m\left(t^{\prime}\right)\\
&+\lr(t)\int_{0}^{t}\lr(t^\prime) R\left(t, t^{\prime \prime}\right) Q^{\prime \prime}\left(C\left(t, t^{\prime \prime}\right)\right) C\left(t^{\prime}, t^{\prime \prime}\right) d t^{\prime \prime} \\
&+\lr(t)\int_{0}^{t^{\prime}}\lr(t^\prime) R\left(t^{\prime}, t^{\prime \prime}\right) Q^{\prime}\left(C\left(t, t^{\prime \prime}\right)\right) d t^{\prime \prime}+ 2T\lr(t)^2 R(t,t^\prime) \\
\frac{\partial}{\partial t} R\left(t, t^{\prime}\right)=&-\spherical(t)\lr(t) R\left(t, t^{\prime}\right) + \delta(t-t^\prime) \lr(t)\\
&+\lr(t)\int_{t^{\prime}}^{t}\lr(t^\prime) R\left(t, t^{\prime \prime}\right) Q^{\prime \prime}\left(C\left(t, t^{\prime \prime}\right)\right) R\left(t^{\prime \prime}, t^{\prime}\right) d t^{\prime \prime} \\
\frac{d}{d t} m(t)=&-\lr(t)\spherical(t) m(t)+\lr(t)Q^{\prime}(m(t))\\
&+\lr(t)\int_{0}^{t} \lr(t^\prime)R\left(t, t^{\prime \prime}\right) m\left(t^{\prime \prime}\right) Q^{\prime \prime}\left(C\left(t, t^{\prime \prime}\right)\right) d t^{\prime \prime} \\
\spherical(t)=&\lr(t)T+Q^{\prime}(m(t)) m(t)\\
&+\int_{0}^{t}\lr(t^\prime) R\left(t, t^{\prime \prime}\right)\left[Q^{\prime}\left(C\left(t, t^{\prime \prime}\right)\right)+Q^{\prime \prime}\left(C\left(t, t^{\prime \prime}\right)\right) C\left(t, t^{\prime \prime}\right)\right] d t^{\prime \prime}
    \end{split}
\end{tcolorbox}
where we defined $Q(\spin)=Q_p(\spin)+Q_2(\spin)=\frac{\spin^{p}}{p \Delta_{p}}+\frac{\spin^{2}}{2 \Delta_{2}}$. The loss is related to $z(t)$ via:
\begin{align}
    z(t) = T\lr(t) - p \frac{\loss_p}{N}- 2 \frac{\loss_2}{N},
\end{align}
with $\loss_2$, respectively $\loss_2$ are the loss associated with the matrix, respectively tensor, channel.
\paragraph{The Langevin easy phase} As explained in \cite{mannelli2020marvels}, one finds different phases in the two dimensional space spamed by the noise intensities $\Delta_2$ and $\Delta_p$. In the \emph{Langevin easy} phase, a system initialised with a magnetisation $m\sim O(\nicefrac{1}{\sqrt N})$ recovers the signal and converges to an overlap of order 1. It is delimited by $\Delta_2<\Delta_2^*$, where $\Delta_2^*$ is the solution to the implicit equation:
\begin{equation}
    \Delta_2<\Delta^\star_2=\sqrt{ \frac{\Delta_p }{ (p-1)  (1-\Delta_2^{*})^{p-3}  }}.
\end{equation}
In contrast, in the \emph{Langevin hard} and \emph{Langevin impossible} phase, i.e. $\Delta_2>\Delta_2^*$, the dynamics fail to recover the signal and remain at low magnetisation. More details in \cite{mannelli2019afraid}.
\subsection{Derivation of the Ground-state Loss}
\label{app:ground_state_loss}
In order to derive the ground state properties of the system, we resort to the replica method, developed in physics as a tool to deal with random systems. Using these tools, involves performing a mapping between the optimisation problem, an inference problem and a physical system. We can consider the estimator $\spin$ as a guess on the planted signal $\signal$ and $y$ be the observations.The, using Bayes formula we can express the posterior probability of the estimator $\spin$ given the observation $y$:
\begin{align}
\label{eq:bayesposterior}
    \begin{split}
        P[x\vert y] = \frac{1}{P[y]} P[x]P[y\vert x] \approx_{\beta=1} \frac{1}{P[y]}P[x]P[y\vert x]^{-\beta} = \frac{1}{Z(y)}\mathrm{e}^{-\beta \loss}.
    \end{split}
\end{align}
We can identify the last terms with a Gibbs distribution at temperature $\beta=\nicefrac{1}{T}$ and $Z$ is a normalisation constant named the \emph{partition function}. At $\beta=1$, the posterior \ref{eq:bayesposterior} is the exact posterior of the problem. At $\beta\to\infty$, the distribution is dominated by the spin configuration minimising the loss, i.e. the maximum likely hood approximator of the problem. 
The partition function, and its logarithm the \emph{free energy}:
\begin{equation}
    \Phi = \frac{- 1}{N}\log Z,
\end{equation}
act as a generating functional. I.e. they encapsulate all the relevant information needed to describe of the system. Notably, all observables can be obtained by taking derivatives of it. In particular, the loss and the overlap with the signal are given by:
\begin{align}
    \begin{split}
        \loss &= \frac{1}{N}\frac{1}{Z}\int_{\S^{N-1}} \loss e^{-\beta \loss} = \frac{-1}{N}\frac{ \partial \log Z}{\partial \beta} = \frac{ \partial \Phi}{\partial \beta}\\
        m & = \frac{1}{N}\sum_{i=1}^N \frac{1}{Z}\int_{\S^{N-1}} \spin_i\signal_i e^{-\beta \loss + \bs{h}\cdot\spin}\vert_{\bs{h} = 0} = - \signal\cdot \grad_{\bs{h}} \Phi .
        \label{eq:loss_deriv}
    \end{split}
\end{align}
The spiked tensor model is rendered more complex due to the randomness associated with the couplings. We need to evaluate the averaged logarithm of the partition function $\overline{\log Z}$ which is in general prohibitive. To deal with this problem, physics have developed the heuristic \emph{replica method} based on the equality:
\begin{equation}
    \overline{\log Z}=\lim_{n\to 0}\frac{\overline{Z^n}-1}{n}.
\end{equation}
In practice, one computes $\overline{Z^n}$ for $n\in\N$ and then extends the result to real $n$. The problem can be viewed as introducing $n$ identical, replicated, copies of the system. As we will see, averaging over the random couplings introduces correlation between the copies. 
$\overline{Z^n}$ can easily be evaluated as:
\begin{align}
    \begin{split}
\overline{\mathcal{Z}^{n}} &=\mathop{\EE}_{\substack{\coupling_{i_1\dots i_p}\\\coupling_{i j}}} \mathop{\int} \prod_{a=1}^{n}\mathrm{e}^{ 
\beta \sqrt{\frac{1}{p \Delta_p N} }\mathop{\sum}_{i_1,\dots,i_p} \coupling_{i_1,\dots,i_p} x_{i_1}^{(a)}\dots x_{i_p}^{(a)}
+ \beta \sqrt{\frac{1}{2 \Delta_2 N} }\mathop{\sum}_{i, j} \coupling_{i, j} x_{i}^{(a)} x_{j}^{(a)}
+N \beta \mathop{\sum}_{i} Q\left( \frac{x_{i}^{(a)} x_{i}^{*}}{N} \right)} \prod_{a=1}^{n} d x^{(a)} 
\\
&=\int_{\mathbb{S}^{n(N-1)}(\sqrt{N})} \mathrm{e}^{ N \beta \mathop{\sum}_{i} Q\left( \frac{x_{i}^{(a)} x_{i}^{*}}{N} \right) + 
 \frac{N \beta^{2}}{2} Q\left(\mathop{\sum}_{a, b=1}^{n}\mathop{\sum}_{i} \frac{x_{i}^{(a)} x_{i}^{(b)}}{N} \right) } \prod_{a=1}^{n} d x^{(a)} .
    \end{split}
\end{align}
where we introduced $Q(x)=\nicefrac{x^2}{2\Delta_2}+\nicefrac{x^p}{p\Delta_p}$.
The second term in the exponent carries the interaction between the different copies obtained after averaging out the random couplings. It depends on the overlap $\bQ$ having entries $\Q_{ab} = \sum_{i} \frac{ x_{i}^{(a)} x_{i}^{(b)} }{N}$. We associate the index $a=0$ with the ground truth signal $\signal$. Using the exponential representation of the Dirac delta function, we introduce the overlap matrix into the partition function. After some manipulation we obtain:
\begin{align}
    \overline{Z^n} &= \int \mathrm{e}^{N \beta S(\bQ) } \\
     \beta S(\bQ)  &= \frac{1}{2} \log \det \bQ +\beta^2 \sum_{a, b=1}^n  Q(\Q_{ab}) + \beta \sum_{a =1}^n  Q(\Q_{a0}).
\label{eq:eqSN}
\end{align}
The factor $N$ in the exponential in the integrand, implies that in the $N\to\infty$ limit, the integral is dominated by the matrix $\Q$ maximising the action $S$. In order to progress, we make a replica symmetric ansatz\footnote{Since we only consider the Langevin easy phase, where there is no ergodicity breaking, we do not need to consider a 1RSB ansatz.}: i.e. we assume the different systems have overlaps $q$ between each other and $m$ with the ground truth. This imposes a matrix $\Q$ has the form:
\begin{align}
    \begin{split}
\Q=\left(\begin{array}{cccc}
1 & m & m & m \\
m & 1 & q & q \\
m & q & 1 & q \\
m & q & q & 1
\end{array}\right).
    \end{split}
\end{align}
Replacing this overlap matrix in \ref{eq:eqSN} and taking $n\to 0$, we obtain:
\begin{align}
    \begin{split}
        \beta S_{\mathrm{RS}}(q, m) = 
        n\left\{ \frac{1}{2} \frac{q-m^{2}}{1-q}+\frac{1}{2} \log (1-q) +\frac{\beta^2}{2} Q(1) - \frac{\beta^2}{2} Q(q) + \beta Q(m) \right\}
    \end{split}
\end{align}
We now maximise $S$ with respect to  $m$ and $q$ and obtain the saddle point equations:
\begin{align}
    \frac{S_{\mathrm{RS}}(q, m)}{\partial m} &= \frac{-m}{1-q} + \beta Q^\prime(m) \\
    \frac{S_{\mathrm{RS}}(q, m)}{\partial q} &= \frac{q-m^2}{(q-1)^2}+\beta ^2 Q^\prime(q) 
    \label{eq:FPRS}
\end{align}

The expression of the loss as a function of the overlaps $m$ and $q$ is given by using Eq.~\ref{eq:loss_deriv}:
\begin{align}
    \begin{split}
        \loss(m, q) = - \beta (Q(1) - Q(q)) + Q(m) 
    \end{split}
    \label{eq:lossRF}
\end{align}
By evaluating the above at the solutions Eqs.~\ref{eq:FPRS}, we obtain the ground state loss at a given temperature.
\paragraph{$T=1$ solution}
At $T=1$ (i.e. $\beta=1$) the posterior Eq.~\ref{eq:bayesposterior} is exact and we can use the Nishimori identity stating that the distribution of the estimator is the same as the one of the signal implying $m=q$. Replacing the identity in Eq.~\ref{eq:FPRS} and in Eq.~\ref{eq:lossRF} we have:
\begin{align}
    \begin{split}
        m = (1-m)Q^\prime(m), 
        \loss^{T=1}_\gs = - Q(1).
    \end{split}
\end{align}

\paragraph{$T=0$ solution}
We can think of the $0$ temperature system (i.e. $\beta\to\infty$) as physical system coupled to a thermal bath. As the temperature goes to $0$, all particles collapse to a point at the minimum of the loss. Thus, the overlap tends to $1$. However, we check that Eqs.~\ref{eq:FPRS} are singular at $q=1$. To properly take the limit, we perform a linear expansion in the temperature by replacing $q=1-\chi T$ in the equations and linearising in $T$. We then obtain the equation for $m$:
\begin{align}
    \begin{split}
   \chi & =  \sqrt{\frac{1-m^2}{Q^\prime(1)}}\\
   m & = \chi Q^\prime(m)
    \end{split}
\end{align}
and the ground state loss:
\begin{align}
    \begin{split}
        \loss^{T\to 0}_\gs & = (-Q(m) - \chi Q^\prime(1)).
    \end{split}
\end{align}

\subsection{Additional results on the optimal learning rate schedule in the SMT model}

\begin{figure}[ht!]
    \centering
    \includegraphics[width=0.9\linewidth]{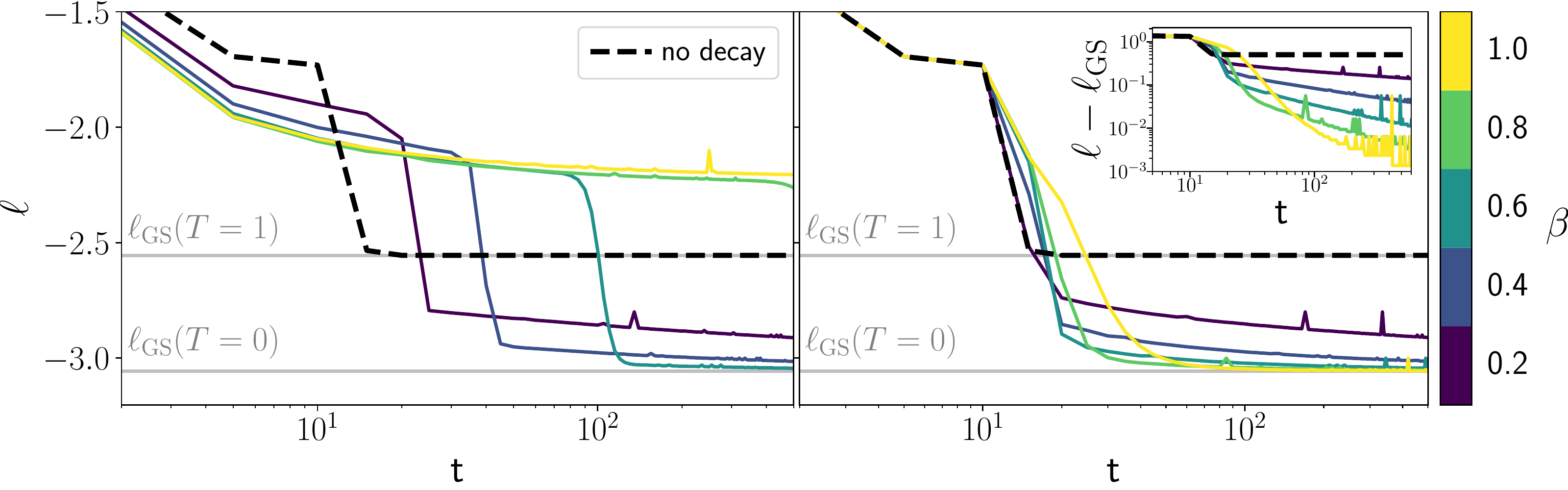}
    \caption{\textbf{Emergence of a crossover time in the SMT model.} By fixing $\lrexponent$ from start, a randomly initialised system will remain stuck at \emph{threshold} states of $0$ overlap until $\tjump$ which is minimal for $\lrexponent=0$. Higher $\lrexponent$, allow to reach lower loss solutions but require much longer to converge. \emph{(Right)} The optimal schedule is to keep $\lr$ constant until $\tjump$ and then setting $\lrexponent=1$. By doing so, we get the best of both worlds: the first phase minimises $\tjump$ while the second allows to reach more informative solutions. \emph{(Inset)} $m-m_\gs$ shows that choosing higher $\lrexponent$ after $\tjump$ allows to reach more informative minima. {\it Parameter: $\lrexponent=0.8$, $\Delta_2=0.2$, $\Delta_p=6$, $\lr_0=1$, $T=1$, $dt = 10^{-2}$, $m_0=10^{-10}$.}}
    \label{fig:app:SMTmodel}
\end{figure}

In this section, we give additional results confirming the optimal decay of the learning rate in the spiked-matrix tensor model. We have seen in the main text, that there is a crossover time $\tjump$ before which the learning rate should be kept fixed as the system is in the search phase. After $\tjump$, the dynamics enter a convex basin and one should decay the learning rate as $\lr(t)\sim t^{-\lrexponent}$. To verify that $\lrexponent=1$ leads to the lowest loss, in the right panel of Fig.~\ref{fig:app:SMTmodel}, we keep a constant learning rate until $t_\mathrm{s}$ after which we vary the exponent with which the learning rate is decayed. We check that $\beta=-1$ allows to reach the best solutions. However, the left panel shows that if the learning rate is decayed from start, the dynamics take much longer to converge towards the signal and remain stuck at high loss for very long.

\subsection{Separation of time scales and matching solution}
\label{sec:matching}
The long time dynamics, i.e. $t\to\infty$ of the $p$-spin model can be separated into two regimes:
\begin{itemize}
    \item For all times $t, t\to\infty$ with $\frac{t-t^\prime}{t}\to 0$ the system is stationary. Here, the dynamics are time-translation invariant (TTI) and the fluctuation-dissipation theorem (FDT) holds. The two time functions $C(t, t^{'})$ and $R(t, t^{'})$ are thus only a function of the time difference $\tau = t-t^{'}$. In this regime, we define $\ctti(\tau)\equiv C(t-t^{'},0)$ and $\rtti(\tau)\equiv R(t-t^{'},0)$. The FDT gives $\rtti(\tau) = -\frac{1}{T}\frac{\dd \ctti(\tau)}{\dd\tau}$. As a consequence, the equations for $R$ and $C$ collapse into a single equation. \\
    \item For all times $t, t\to\infty$ with $\frac{t-t^\prime}{t}= O(1)$ the system \emph{ages} i.e. the dynamics remain trapped in metastable states and does not lose memory of its history. The relevant variable to consider in this regime is $\lambda=\nicefrac{t^\prime}{t}$. The correlation and response functions can be rescaled as $\ra(\lambda)=tR(t, t^\prime)$ and $q\ca(\lambda)=C(t, t^\prime)$ with $q = \lim_{\tau\to\infty}\ctti(\tau)$. In this aging regime, a generalised form of the FDT holds and $\ra(\lambda) = \frac{x}{T}q\frac{\dd \ca(\lambda)}{\dd\lambda}$. The \emph{violation parameter} $\spin$ is found by \emph{matching} i.e. considering the equations for the response and the correlation separetly. $q$ is found by imposing $q = lim_{\tau\to\infty}\ctti(\tau)$ in the equation of the TTI regime.
\end{itemize}
In order to derive analytical results, we use the hypothesis of these two times regimes to split the time integrals in Eqs.~\ref{eq:dynamical_equations_no_signal}. For compactness we also define $Q(x)=\nicefrac{x^p}{2}$. As noted in the main text, we can re-scale time according to $\dd \tilde{t} = \eta(t)\dd t$ and obtain a system at an effective temperature $\tilde{T} =  (1-\lrexponent)^{\frac{1}{1-\lrexponent}}\frac{T}{ t^{\nicefrac{\lrexponent}{1-\lrexponent}} }$. We are ultimately interested in determining the threshold loss, a static quantity, and can hence perform its derivation using a constant learning rate. This analysis is a special case of the more general on performed in \cite{mannelli2019afraid}. Here, we show it for the special case of the $p$-spin model with no signal. In particular, we skip all the computations and refer the reader to \cite{mannelli2019afraid} (Appendix B) for additional details.

\paragraph{Lagrange multiplier in the long time-limit}
Let us start to illustrate how to proceed by computing the long time limit of the loss $\muinf= \lim_{t\to\infty}\spherical(t)$ using Eqs.~\ref{eq:dynamical_equations_no_signal}:
\begin{align}
    \begin{split}
    \label{app:mu_infty1}
        \muinf(T) - T &=  p\underbrace{\int_{0}^t dt^{\prime\prime}  R(t^{\prime\prime}, t) Q^{\prime}(C(t^{\prime\prime}, t))}_{\int_{\mathrm{TTI}} + \int_{\mathrm{aging}}}\\
        &=-\int_{0}^{\infty} \frac{1}{T} \frac{d}{d \tilde{t}} Q\left(C_{\mathrm{TTI}}(\tilde{t})\right) d \tilde{t}+\int_{0}^{1} \mathcal{R}(\lambda) Q^{\prime}(q \mathcal{C}(\lambda)) d \lambda \\
        \Leftrightarrow\muinf&=\frac{1-q^{p}}{2 T}+\int_{0}^{1} \mathcal{R}(\lambda) Q^{\prime}(q \mathcal{C}(\lambda)) d \lambda, 
    \end{split}
\end{align}
where we used the fact that by definition $\ctti(\infty)=q$ and $\ctti(0)=1$. Also note that we neglected all the finite time contribution to the integrals. We are going to determine $\muinf$ using this equation.

\paragraph{Stationary regime}
In order to find the dynamical equations in the stationary regime, we proceed as before and separate the contributions of the TTI regime from those of the aging regime in the integrals. Since both equations for the response and the correlation collapse into a single equation, we consider only the evolution of the correlation $\ctti$. Using Eqs.~\ref{eq:dynamical_equations_no_signal} we have:
\begin{align}
    \begin{split}
        (\muinf + \partial_\tau)\ctti(\tau) =  \int_{0}^{t_{1}} d t^{\prime \prime} R\left(t_{1}, t^{\prime \prime}\right) Q^{\prime\prime}\left(C\left(t_{1}, t^{\prime \prime}\right)\right) C\left(t^{\prime \prime}, t_{2}\right) + \int_{0}^{t_2} dt^{\prime\prime}  R(t^{\prime\prime}, t_2) Q^{\prime}\left(C(t^{\prime\prime}, t_1)\right)
    \end{split}
\end{align}
Using Eqs. 62 of \cite{mannelli2019afraid}, we have: 
\begin{equation}
    \partial_{\tau} C_{\mathrm{TTI}}(\tau)+\left(\frac{1}{T} Q^{\prime}(1)-\mu_{\infty}\right)\left[1-C_{\mathrm{TTI}}(\tau)\right]+T=-\frac{1}{T} \int_{0}^{\tau} Q^{\prime}\left(C_{\mathrm{TTI}}\left(\tau-\tau^{\prime \prime}\right)\right) \frac{d}{d \tau^{\prime \prime}} C_{\mathrm{TTI}}\left(\tau^{\prime \prime}\right) d \tau^{\prime \prime}
\end{equation}
When $\tau\to\infty$, the time variations of  $C_{\mathrm{TTI}}(\tau)$ are negligible. Taking this limit in the above equation gives:
\begin{equation}
    \muinf =\sqrt{Q^{\prime\prime}(q)}+\frac{Q^{\prime}(1)-Q^{\prime}(q)}{T}
\end{equation}
This equation allows to determine the threshold loss, i.e. the loss at the plateau reached by the system before the recovery of the signal. We notice that the equality above holds for all $\Delta_2$, $\Delta_p$ and hence also if one of the two is sent to infinity. Therefore, we have:
\begin{equation}
    \ell_\th = \ell_{p} +  \ell_{2},
\end{equation}
with $\ell_{2} = \frac{1}{2}(\lr(t) - \spherical_{\infty;\Delta_{p}\to\infty})$ and similarly for $\ell_p$. 
Thus, by defining $Q_k(x)=\nicefrac{x^k}{k\Delta_k}$, we obtain:
\begin{equation}
    \ell_{k} = \frac{1}{k}\left(\lr(t) - \sqrt{Q_k^{\prime\prime}(q)}-\frac{Q_k^{\prime}(1)-Q_k^{\prime}(q)}{T}\right)
    \label{eq:eth_SMTM}
\end{equation}
Using this equation, and performing an expansion around $0$ for $T$ and $1$ for $q$, we can determine that at low temperatures, the threshold energy scales linearly with $T$.
\newpage

\section{Additional results for the Teacher-Student Regression Task}
\label{app:teacher-student}

\begin{figure}[!ht]
    \centering
        \subfloat[$K=M=5$]{\includegraphics[width=0.45\columnwidth]{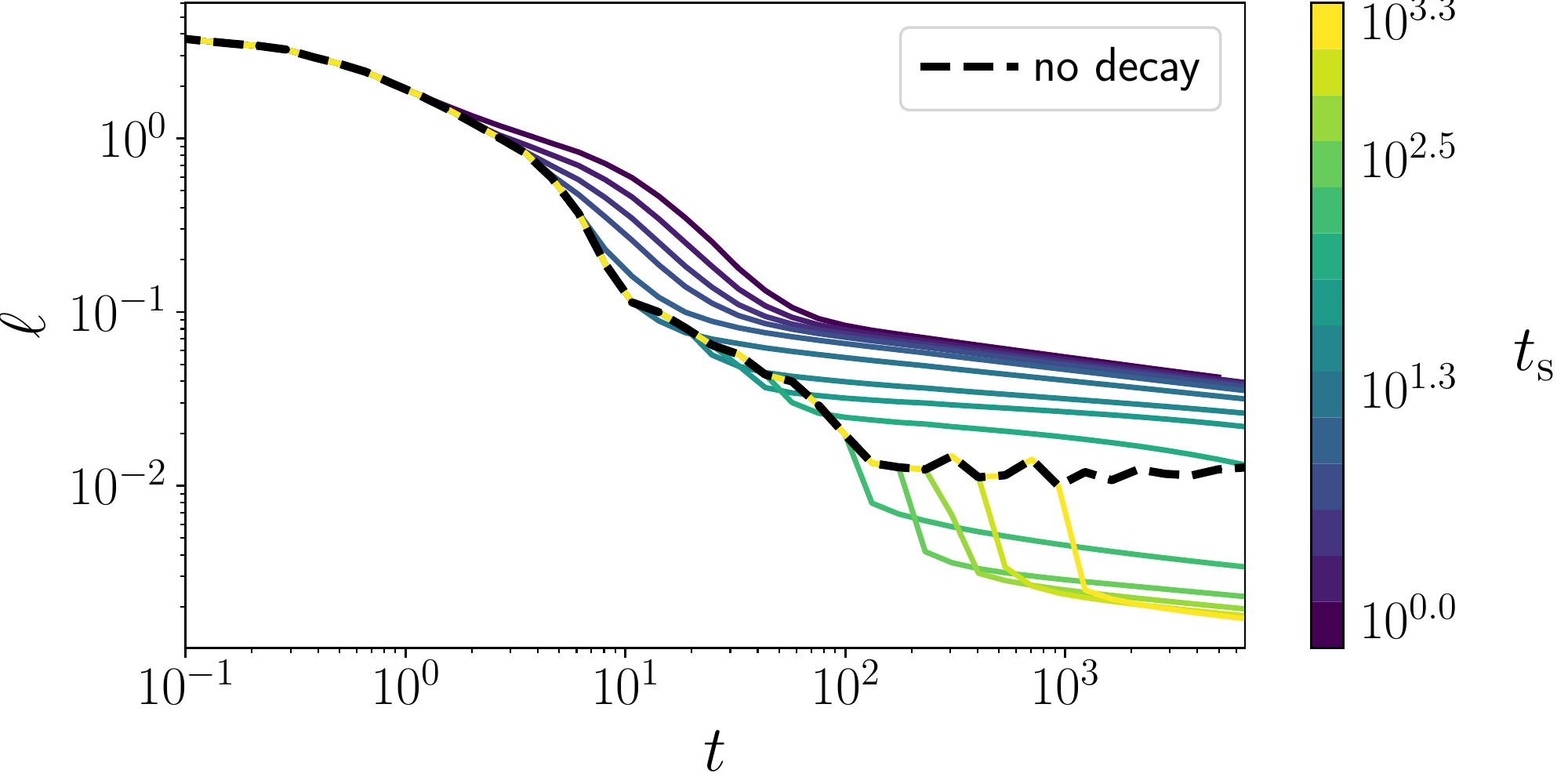}}
        \qquad
        \subfloat[$K=M=20$]{\includegraphics[width=0.45\columnwidth]{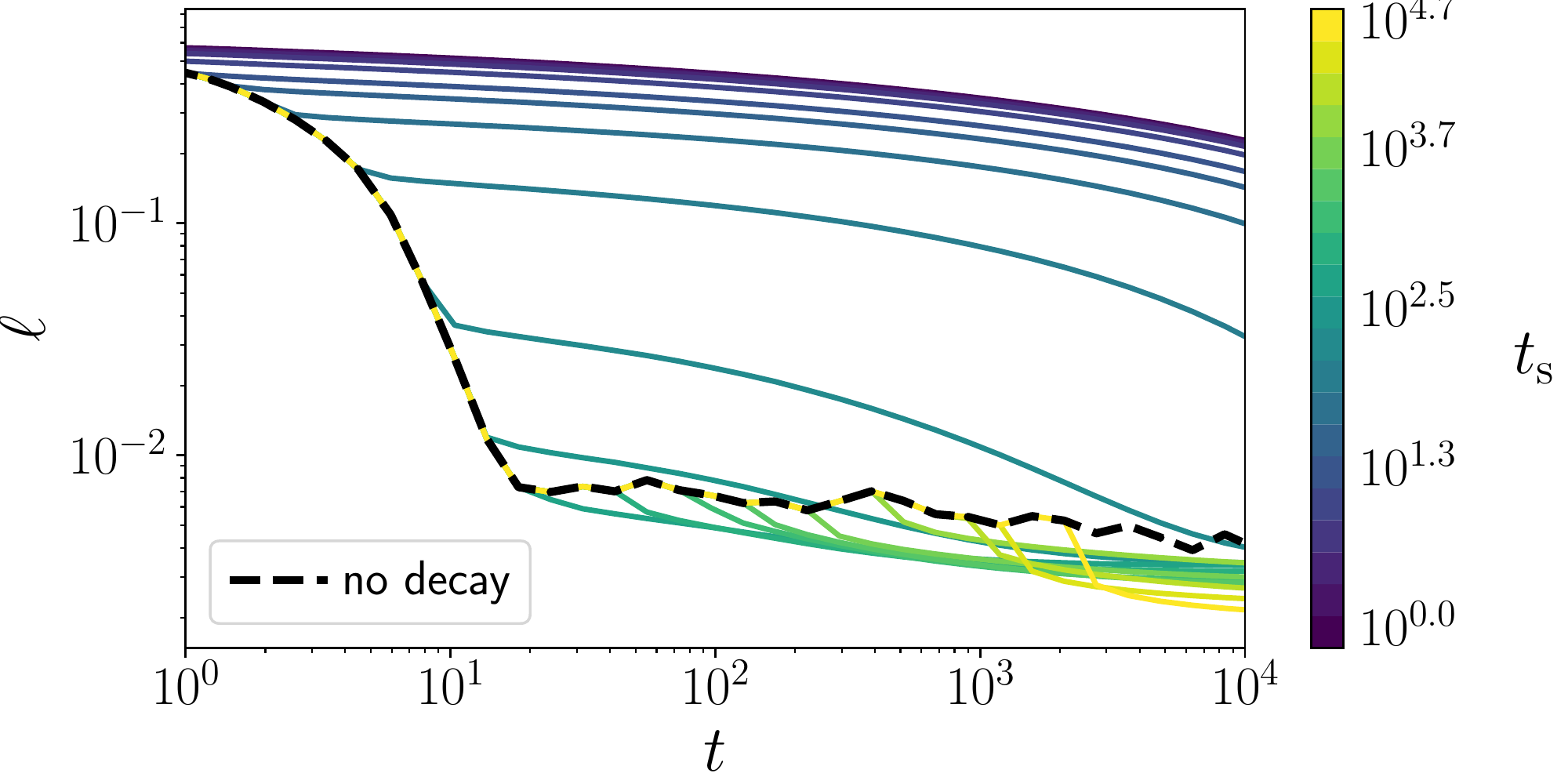}}
        \caption{\textbf{The crossover time is also reflected in a regression task with SGD.} A student with $K$ hidden nodes is trained to reproduce the output of her $M$ hidden nodes. \emph{(Left)} $K=M=5$. \emph{(Right)} $K=M=20$. As in the main text.before, we find that decaying the learning rate before the loss plateaus performance, but decaying as $\lr(t)\sim t^{-1}$ once the plateau is reached allows to reach zero loss. {\it Parameters: $N=500$, $P=10^4$, $\lr_0=10^{-1}$, $\lrexponent=0.8$.}}
    \label{fig:app_app_teacher_student}
\end{figure}

In this section we give additional results on the teacher-student regression task discussed in Sec.~\ref{sec:TS}. The setting is the same as in the main text: a $K$ hidden nodes 2 layer neural network student is trained to reproduce the output of her 2 layer neural network teacher of $M$ nodes on gaussian inputs. We train the model with on a finite dataset of $P$ examples using a mini-batch size $B=1$.
Fig.~\ref{fig:app_app_teacher_student} verifies that the conclusions drawn in the main text hold for different values of $K$ and $M$. The optimal schedule is to keep the learning rate constant until $\tjump$ and to then decay it as $\nicefrac{1}{t}$. If the learning rate is decayed too soon, i.e. at $t_s<\tjump$, learning remains stuck at high loss values. Decaying after $\tjump$ instead allows to reduce the noise in optimisation and reach lower loss solutions. We verify that in both these cases, $\tjump$ matches the end of the "specialisation" transition, where the loss achieved student trained at constant learning rate plateaus.

\end{document}